\PassOptionsToPackage{table}{xcolor}
\documentclass[10pt,twocolumn,letterpaper]{article}

\usepackage[pagenumbers]{cvpr}      
\usepackage{xspace}

\definecolor{cvprblue}{rgb}{0.21,0.49,0.74}
\usepackage[pagebackref,breaklinks,colorlinks,allcolors=cvprblue]{hyperref}
\usepackage{amsmath}
\usepackage{amssymb}
\usepackage{placeins}
\usepackage[T1]{fontenc}
\usepackage{booktabs}
\usepackage{multirow}
\usepackage{xspace}
\usepackage{colortbl}
\usepackage{tabularx}
\newcommand{\mainref}[1]{#1 of the main paper}
\usepackage[table]{xcolor}

\definecolor{oursgray}{gray}{0.94}
\newcommand{\hl}{\cellcolor{oursgray}}
\def\confName{CVPR}
\def\confYear{2027}

\title{BRepJEPA: Predictive Self-Supervised Learning on\\ Heterogeneous B-Rep Graphs}

\author{
Konstantinos Gkrispanis$^{*}$ \quad
Stavros Nousias \quad
Andr\'{e} Borrmann\\[3pt]
Technical University of Munich (TUM)\\
{\small\ttfamily \{konstantinos.gkrispanis,stavros.nousias,andre.borrmann\}@tum.de}
}

\newcommand{\method}{BRepJEPA\xspace}

\begin{document}
\twocolumn[{%
\renewcommand\twocolumn[1][]{#1}%
\maketitle
\vspace{-2.6em}
\begin{center}
  \captionsetup{type=figure, font=footnotesize}
  \includegraphics[width=\textwidth]{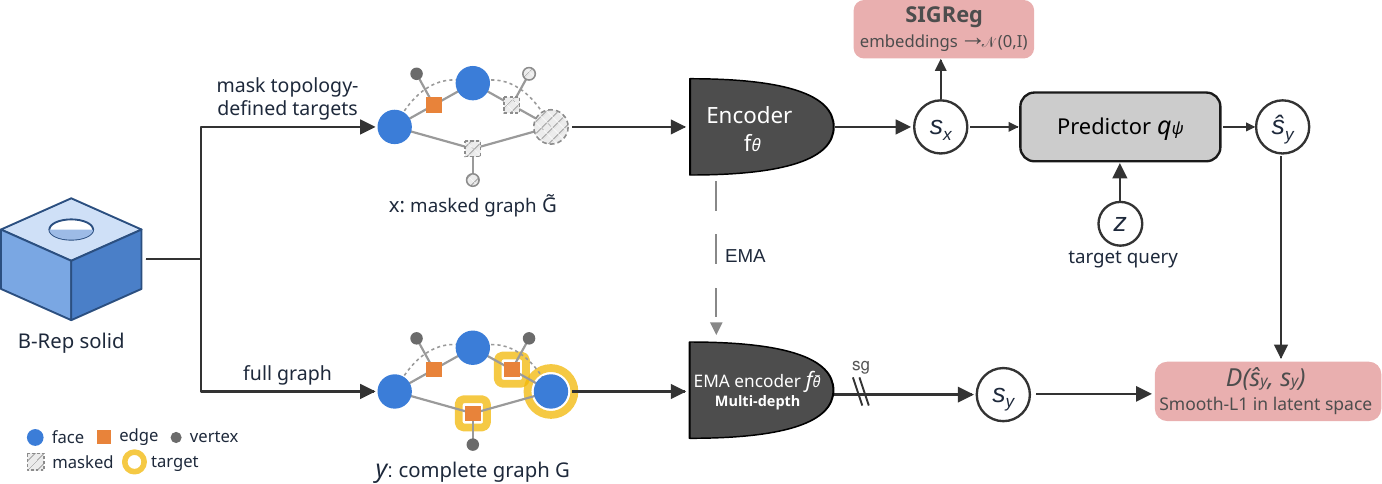}
  \vspace{-0.4em}
  \captionof{figure}{\textbf{\method architecture.}
The context encoder $f_\theta$ processes a masked B-Rep graph ($x$),
while the EMA teacher $f_{\bar\theta}$ encodes the complete graph ($y$).
Conditioned on a target query $z$, the predictor $q_\psi$ predicts
multi-depth teacher representations from the online embeddings $s_x$.
SIGReg regularizes the online embeddings toward an isotropic Gaussian.}
  \label{fig:teaser}
\end{center}
\vspace{0.6em}
}]
\begingroup
\renewcommand{\thefootnote}{*}
\footnotetext{Corresponding author. Code and pretrained models:
\url{https://github.com/gkrispanis/BRepJEPA}.}
\endgroup
\begin{abstract}
 We introduce \method, the first joint-embedding predictive pretraining method for B-Reps. Unlike existing reconstruction-based methods for CAD, \method predicts the latent representations of masked boundary loops, oriented edges, region boundaries, and faces from the rest of the solid. Our targets average normalized states across all layers of an exponential-moving-average teacher, while a sketched isotropic-Gaussian regularizer prevents dimensional collapse. We pretrain on 100k ABC solids and evaluate on four tasks against BRepMAE, Brep2Shape, UV-Net, and Graph-JEPA using their official implementations. \method is the strongest of the evaluated frozen encoders on every task, with the largest margins when labels are scarce. Ablations show that multi-depth targets contribute most of the gain and that topology-defined targets provide additional improvements. These results establish latent prediction as an effective means of pretraining B-Rep encoders.
 \end{abstract}

\vspace*{-2.5em}
\section{Introduction}
\label{sec:intro}

\begin{figure*}[t]
  \centering
  \includegraphics[width=\textwidth]{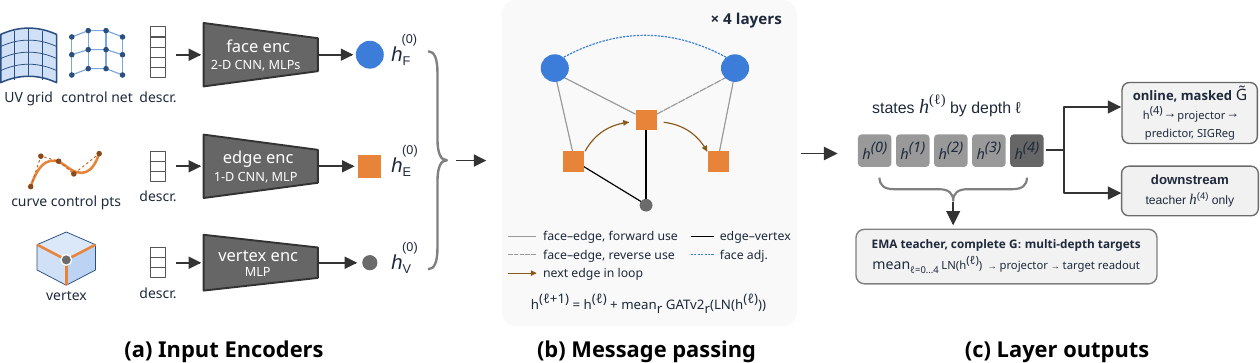}
  \caption{\textbf{The B-Rep encoder} shared by the online branch and the EMA teacher. (a)~ The input encoders map each face, edge, or vertex to a $256$-dimensional token $h^{(0)}$: faces from a $16{\times}16$ UV grid of positions, normals, and trimming mask, the NURBS control net, and $17$ descriptors. Edges from the curve control points and $8$ descriptors. Vertices from $8$ descriptors. In the online branch, a masked entity's token is replaced by a learned [MASK] token. (b)~Four heterogeneous GATv2 layers pass messages over eight typed relations, with one attention per relation, averaged over relations, and pre-LayerNorm residual updates. (c)~The online branch feeds its final state $h^{(4)}$ to the projector, predictor, and SIGReg. The teacher's targets average the LayerNorm-normalized states of all five depths (\cref{eq:multidepth}). Downstream tasks keep only the teacher's final state.}
  \label{fig:arch}
  \vspace{-4pt}
\end{figure*}

Boundary representations encode CAD solids through trimmed parametric surfaces, curves, vertices and their incidence relations. CAD-native networks~\cite{brepnet,uvnet,topognn} have shown the value of operating on this structure directly. Annotating B-Reps is expensive, while large collections of unlabeled STEP models are available~\cite{abc,willis2021fusion360,wu2021deepcad}. Researchers have explored many approaches to exploit these models. Existing approaches largely rely on reconstructing masked geometric or topological attributes~\cite{jones2023ssl,brepmae,hmaebrep,brep2shape}, building on developments in masked autoencoding~\cite{devlin2019bert,mae,hou2022graphmae,pang2022pointmae,yu2022pointbert}. Nevertheless, reconstruction asks the network to reproduce input attributes, not to organize its latent space around meaningful manufacturing structure. Inspired by I-JEPA~\cite{ijepa}, Graph-JEPA~\cite{graphjepa}, and LeJEPA~\cite{balestriero2025lejepa}, we introduce \method, which pretrains B-Rep encoders through prediction in latent space.

Latent-space prediction involves three design choices: what to predict, which context to use, and how to represent the targets. For B-Reps, the topology offers natural candidates for the first, such as the boundary loops of a face, the coedge through which a face uses an edge and the enclosing boundary of a connected face region. \method masks such units and predicts their latent representations from the remaining graph, with a predictor that is told which entities form the target. These units outperform random local units of matched size by a consistent margin ($0.6$--$1.5$ points, \cref{sec:ablations}). The third choice matters most. A target derived from the final layer of the teacher emphasizes contextual information, allowing the encoder to discard local geometric details that are not needed for prediction. We therefore construct targets by averaging normalized representations across all teacher layers, from the initial geometric encoding to the final message-passing layer, to preserve both local geometry and contextual information. A sketched isotropic-Gaussian regularizer (SIGReg)~\cite{balestriero2025lejepa} on the online embeddings prevents dimensional collapse without the variance and covariance heuristics of earlier JEPAs.

We pretrain \method on 100k ABC solids~\cite{abc} and compare it with BRepMAE~\cite{brepmae}, Brep2Shape~\cite{brep2shape}, UV-Net~\cite{uvnet} and Graph-JEPA~\cite{graphjepa}, using their official published implementations and the same pretraining data. We evaluate frozen and fine-tuned encoders on four tasks: face segmentation on Fusion~360 and MFInstSeg, part classification on TMCAD and edit-outcome retrieval on BrepEDIT. Our contributions are:
\begin{itemize}
\item We introduce \method, the first joint-embedding predictive pretraining method for B-Reps. It predicts four families of topology-defined targets on heterogeneous B-Rep graphs with target-identifying queries, multi-depth teacher targets and a SIGReg-regularized objective.
\item We establish that with frozen encoders our method is the strongest of the evaluated methods on all four tasks, against published methods and against other objectives on our encoder. It improves over masked reconstruction by up to $+9.5$ macro-F1 points, with margins that grow as labels become scarce, and it leads on edit-outcome retrieval.
\item We demonstrate through ablations of teacher depth, target design, predictor queries, masking and regularization that multi-depth targets contribute most of the gain and that topology-defined targets add a consistent gain over random local targets of matched size.
\end{itemize}

\begin{figure*}[t]
  \centering
  \includegraphics[width=\textwidth]{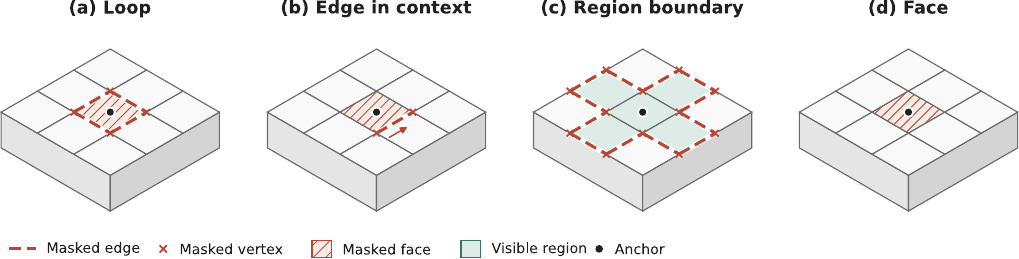}
  \caption{\textbf{Target families} on a heterogeneous B-Rep graph: (a)~boundary loop, (b)~orientation-aware coedge, (c)~enclosing boundary of a visible connected face region, without internal shared edges and (d)~masked face. Training combines the masks shown separately here. The complete-graph teacher supplies all target latents. $\rho$ averages a set's edge latents and $g$ is the fixed orientation-conditioned coedge map.}
  \label{fig:targets}
  \vspace{-4pt}
\end{figure*}
\section{Related Work}
\label{sec:related}

\paragraph{Networks on B-Reps.}
CAD-native architectures that operate on the B-Rep directly rather than a tessellated approximation have shown strong performance. BRepNet~\cite{brepnet} defines topological convolutions over coedges, UV-Net~\cite{uvnet} couples UV-grid CNN features with face-adjacency message passing and BRT~\cite{zou2025brt} brings transformer encoders to B-Rep learning. TopoGNN~\cite{topognn}, whose heterogeneous face/edge/vertex graph and NURBS-based geometric inputs we adopt, addresses industrial machining-feature recognition~\cite{lim2023machining,fuchs2026featurefox}. Multi-task GNNs introduced alongside the MFTRCAD and MFInstSeg datasets~\cite{mftrcad,mfinstseg} also work on that downstream task. Recent work shows that such networks can be sensitive to how a solid is partitioned into faces~\cite{jiang2026canonical} and to its orientation~\cite{ballegeer2026fovnet}. B-repLer~\cite{liu2026brepler} edits B-Reps from language and releases the edit pairs on which we build an edit-outcome retrieval task.

\paragraph{Self-supervised learning on B-Reps.}
Most existing B-Rep pretraining reconstructs geometric and topological attributes. BRepMAE~\cite{brepmae} and hierarchical masked autoencoding~\cite{hmaebrep} recover attributes of masked entities, Jones~\etal~\cite{jones2023ssl} learn face embeddings through neural-field reconstruction and Brep2Shape~\cite{brep2shape} regresses evaluated surface and curve points from parametric primitives. BRep-BERT~\cite{brepbert} combines masked entity modeling with contrastive objectives, while BRepCLIP~\cite{brepclip} aligns B-Reps with text and images. Masked topology modeling~\cite{jiang2026mtm} uses kernel-computed topological attributes for supervision and CAD-kernel edits to construct contrastive views. Unlike reconstruction-based approaches, \method predicts latent representations of masked B-Rep structures. We compare against existing methods using their official released implementations on the same pretraining dataset and evaluate reconstruction-based and predictive objectives on a shared encoder.

\paragraph{Predictive latent self-supervision.}
Joint-embedding predictive architectures (JEPAs)~\cite{ijepa,bardes2024vjepa,garrido2024iwm,ghaemi2025seqjepa,saito2025pointjepa,zhang2026nodejepa} have emerged as an alternative to reconstruction-based self-supervised learning. I-JEPA~\cite{ijepa} predicts representations of masked image regions from positional mask tokens, Graph-JEPA~\cite{graphjepa} predicts embeddings of graph partitions, and HP-JEPA~\cite{xu2026hpjepa} predicts targets at multiple partition resolutions. data2vec~\cite{data2vec,baevski2023data2vec2} regresses the average of normalized teacher layers, which we adapt to our heterogeneous B-Rep entities. Representation collapse~\cite{jing2022dimcollapse} is prevented by variance--covariance regularization~\cite{vicreg,zbontar2021barlow} or, in LeJEPA~\cite{balestriero2025lejepa}, by matching embeddings to an isotropic Gaussian with a sketched statistic (SIGReg). Whether semantically defined prediction units outperform random units of matched statistics has been examined in language~\cite{joshi-etal-2020-spanbert}, on graphs~\cite{graphjepa,wang2026biomjepa}, and in image JEPAs~\cite{xie2026jeparecoverable}. We investigate this choice for B-Rep graphs by comparing topology-defined targets with random targets of matched size.
\section{Method}
\label{sec:method}

\subsection{Heterogeneous B-Rep Graph}
\label{sec:graph}
We follow the graph modeling of TopoGNN~\cite{topognn} and represent each solid as a heterogeneous graph with face, edge and vertex nodes. Eight directed relation types describe face adjacency, face--edge incidence in both directions, edge ordering within loops (\texttt{next\_in\_loop}), and edge--vertex incidence in both directions. Type-specific encoders process sampled UV grids, NURBS control nets and handcrafted descriptors. The face encoder consists of: (i)~a CNN over the sampled UV grid, (ii)~a point network with masked mean pooling over the control net and (iii)~a descriptor MLP for the handcrafted features. The edge encoder combines a 1D CNN over control points and a descriptor MLP and the vertex encoder uses an MLP. A four-layer heterogeneous GATv2~\cite{velickovic2018gat,brody2022gatv2} network with relation-specific convolutions~\cite{schlichtkrull2018rgcn} produces 256-dimensional entity embeddings. The complete B-Rep encoder is shown in \cref{fig:arch}.
We denote the initial embedding of entity $u$ by $h_u^{(0)}$ and its representation after message-passing layer $\ell$ by $h_u^{(\ell)}$, where $\ell\in\{1,\dots,L\}$ and $L=4$. Each layer uses pre-normalization~\cite{xiong2020prenorm} and a residual connection:
\[
h^{(\ell)}=h^{(\ell-1)}+\operatorname{Dropout}\bigl(\operatorname{Conv}_{\ell}(\operatorname{LN}(h^{(\ell-1)}))\bigr),
\]
where $\operatorname{Conv}_{\ell}$ applies a GATv2 convolution to each relation type, with eight attention heads of dimension $32$ and averages the resulting representations across relations. The encoder has $4.99$\,M parameters.

\subsection{Topology-Defined Targets}
\label{sec:targets}

\Cref{fig:targets} summarizes the four target families:
\begin{itemize}
    \item \textbf{Loop} masks all edges of a sampled boundary loop and predicts their pooled latent representation.
    \item \textbf{Edge-in-context} masks one coedge (an edge associated with a specific face and orientation), along with its end vertices and the features of the incident face. The target is conditioned on the coedge orientation.
    \item \textbf{Region-boundary} keeps a sampled connected face region visible and predicts the pooled latent representation of its masked edges.
    \item \textbf{Face} masks one face and predicts its latent representation from adjacent structure.
\end{itemize}

Three implementation details are worth noting. (i)~Loop and region-boundary targets contain variable-size sets of edges, whose latent representations are averaged. A randomly initialized, frozen attention-based readout performed comparably in a preliminary single-seed experiment ($+0.001$ macro-F1, supplementary material), so we use mean pooling. (ii)~The coedge target uses a fixed orientation-conditioned map,
$z_s = g([\,z_e;\,z_f;\,e_o\,])$,
where $z_e$ and $z_f$ are the teacher representations of the edge and its incident face and $e_o$ is an orientation embedding. The map $g$ is randomly initialized and remains frozen, with no optimizer or EMA updates. This introduces orientation dependence without adding trainable parameters to the teacher branch. (iii)~Vertices incident to masked edges are also masked to prevent direct access to endpoint coordinates.

For each graph, we sample two loop targets, four coedge targets, one region-boundary target ($1$--$3$ face-adjacency hops, $3$--$25$ faces) and three face targets, resampling them at every step. Loops are sampled uniformly from face loops with at least three edges, excluding those that share edges with previously selected loops. Coedges are sampled uniformly from face--edge incidences, with each edge selected at most once per solid. Regions are grown from uniformly sampled seed faces and accepted only if they do not cover the entire solid and contain at least three boundary edges not used by earlier region targets. Face targets are sampled uniformly from faces with at least two adjacent faces.

The online view $\widetilde{G}$ combines the masks from all target families. Masked entities receive learned type-specific mask tokens instead of their original features. Incidence and adjacency relations remain visible, while loop-succession relations involving masked edges are removed. On ABC, each solid provides an average of $9.9$ targets, with $21.7\%$ of faces and $22.0\%$ of edges masked per batch.

\subsection{Predictive Architecture}
\label{sec:arch}
We adopt the teacher--student architecture of I-JEPA~\cite{ijepa}. The online branch encodes the masked graph $\widetilde{G}$ and applies a projector and a predictor (\cref{fig:teaser,fig:arch}). An EMA teacher~\cite{tarvainen2017meanteacher,grill2020byol,caron2021dino} encodes the unmasked graph $G$ and provides detached prediction targets.

\paragraph{Multi-depth teacher targets.}
For each face and edge, we average the teacher's normalized representations across all layers:
\begin{equation}
  \bar{h}_u=\frac{1}{L+1}\sum_{\ell=0}^{L}\operatorname{LN}\!\bigl(h_{u,\bar\theta}^{(\ell)}\bigr),
  \label{eq:multidepth}
\end{equation}
where $\operatorname{LN}$ is a parameter-free layer normalization~\cite{ba2016layernorm} across channels and $\bar\theta$ denotes the EMA parameters. The teacher projector maps $\bar{h}_u$ to $\bar z_u$, which is used to construct the targets described in \cref{sec:targets}. The initial representation $h^{(0)}$ captures local geometry, while deeper layers incorporate information from neighboring entities. Averaging these representations provides targets that contain both local and contextual information. Normalization prevents layers with larger activation from dominating the average~\cite{data2vec}. The student uses its final-layer representation, as do downstream tasks evaluated with the teacher encoder. Multi-depth target construction therefore introduces no additional parameters or inference cost.

\begin{figure*}[t]
  \centering
  \includegraphics[width=\textwidth]{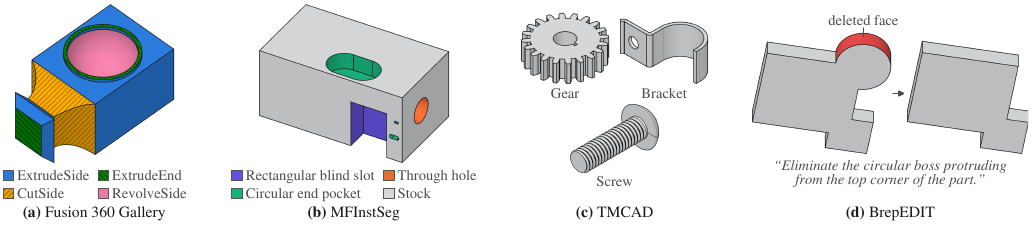}
  \caption{\textbf{Downstream tasks.} (a)~Fusion~360 Gallery face segmentation into eight modeling operations. (b)~MFInstSeg machining-feature face segmentation (25 classes). (c)~TMCAD part classification (ten classes). (d)~BrepEDIT edit-outcome retrieval from a natural-language instruction.}
  \label{fig:benchmarks}
  \vspace{-4pt}
\end{figure*}

\paragraph{Target-conditioned prediction.}
Each target is sampled at an anchor face (shown as a dot in \cref{fig:targets}). The predictor $q_\psi$ is a two-layer pre-norm transformer that takes as input a learned \textsc{cls} token, the anchor-face embedding, the target family and orientation embeddings ($\mathrm{fwd}/\mathrm{rev}/\mathrm{n\text{-}a}$) and one token per target entity. Entity tokens are obtained from the projected online embeddings of the target's masked edges or face, with an additional embedding that indicates the entity type. Similar to the positional mask tokens in I-JEPA~\cite{ijepa}, we replace the original input features with mask tokens, so that these representations encode the target's position within the graph rather than its original geometry. They also distinguish targets sharing the same anchor face, family, and orientation, such as two loops of the same face, which would otherwise receive identical predictions. Targets containing more than $64$ entities are represented by $64$ evenly spaced entity tokens. Finally, a LayerNorm followed by a two-layer MLP maps the transformer output at the \textsc{cls} position to the predicted latent $\widehat{z}_s$.

\paragraph{Objective.}
Let $\mathcal{S}_\tau$ contain the targets of family $\tau$ in a batch and $\mathcal{T}$ the families present. We average the prediction loss within each family and then across families, preventing more frequent target families from dominating the objective and maintaining consistent normalization when families are removed in ablations:
\begin{equation}
  \mathcal{L}_{\mathrm{pred}}= \frac{1}{|\mathcal{T}|}\sum_{\tau\in\mathcal{T}}\frac{1}{|\mathcal{S}_\tau|}\sum_{s\in\mathcal{S}_\tau}\operatorname{SmoothL1}\bigl(\widehat{z}_s,\operatorname{sg}[z_s]\bigr),
  \label{eq:pred}
\end{equation}
where $\operatorname{sg}$ denotes stop-gradient. We apply SIGReg~\cite{balestriero2025lejepa} to the online projected face and edge embeddings of the masked view to prevent dimensional collapse. SIGReg projects the embeddings onto $M$ random unit directions and penalizes the Epps--Pulley~\cite{epps1983normality} distance (along each direction) between the empirical characteristic function and that of a standard Gaussian. For a set $Z$ of embeddings and directions $a_1,\dots,a_M$,
\begin{equation}
  \mathrm{SIGReg}(Z)=\frac{|Z|}{M}\sum_{m=1}^{M}\int_{-3}^{3}\bigl|\hat\varphi_m(t)-e^{-t^2/2}\bigr|^2e^{-t^2/2}\,dt,
  \label{eq:sigreg}
\end{equation}
where $\hat\varphi_m(t)=\frac{1}{|Z|}\sum_{z\in Z}e^{\mathrm{i}t\,a_m^{\top}z}$ is the empirical characteristic function of the projections onto direction $a_m$. The integral is evaluated by the trapezoidal rule on $17$ points of $t\in[0,3]$, using its symmetry in $t$. Because the faces of one solid are not independent samples, SIGReg is evaluated on sub-batches that contain one entity per solid. A batch of $N$ entities from $B$ solids gives $\lceil N/B\rceil$ such sub-batches, each with one randomly drawn entity of every solid, and \cref{eq:sigreg} is averaged over them. The training loss is:
\begin{equation}
  \mathcal{L}=(1-\lambda)\,\mathcal{L}_{\mathrm{pred}}+\lambda\,\tfrac12\bigl[\mathrm{SIGReg}(Z_F)+\mathrm{SIGReg}(Z_E)\bigr],
  \label{eq:loss}
\end{equation}
with $\lambda{=}0.05$ and $M{=}1{,}024$ directions re-drawn at every step. After each step, EMA updates the teacher encoder and projector, $\bar\theta\leftarrow\mu\bar\theta+(1-\mu)\theta$, with $\mu$ increasing from $0.996$ to $1$ on a cosine schedule. An alternative approach to prevent representation collapse is the variance--covariance regularization. We compare these two approaches in \cref{sec:ablations}. After pretraining, only the teacher encoder is retained for downstream evaluation. 
\section{Experiments}
\label{sec:experiments}

\subsection{Experimental Setup}
\label{sec:setup}

\begin{table*}[t]
  \centering
  \caption{\textbf{Frozen-encoder transfer.}
    Comparison of self-supervised pretraining methods on five downstream metrics after pretraining on 100k ABC solids. Macro-F1 (\%) for Fusion~360, MFInstSeg, and TMCAD. MRR and R@1 ($\times100$) for BrepEDIT. Brep2Shape and UV-Net use one seed.}
  \label{tab:comparison}
  \vspace{-4pt}
  \footnotesize
  \setlength{\tabcolsep}{5pt}
  \setlength{\aboverulesep}{0pt}\setlength{\belowrulesep}{0pt}\renewcommand{\arraystretch}{1.12}
  \begin{tabularx}{0.95\textwidth}{@{}l*{5}{>{\centering\arraybackslash}X}@{}}
    \toprule
    & \multicolumn{3}{c}{Segmentation and classification} & \multicolumn{2}{c}{Edit retrieval} \\
    \cmidrule(lr){2-4}\cmidrule(lr){5-6}
    Method & Fusion~360 & MFInstSeg & TMCAD & MRR & R@1 \\
    \midrule
    Random initialization (reference) & $75.85{\pm}1.57$ & $79.76{\pm}0.86$ & $77.33{\pm}0.95$ & $65.4{\pm}0.4$ & $45.4{\pm}0.5$ \\
    \midrule
    UV-Net~\cite{uvnet}& $35.05$ & $21.46$ & $69.48$ & $61.6$ & $39.9$ \\
    Graph-JEPA~\cite{graphjepa} & $57.45{\pm}6.16$ & $66.69{\pm}7.00$ & $73.55{\pm}3.27$ & $61.8{\pm}0.8$ & $40.3{\pm}1.1$ \\
    Brep2Shape~\cite{brep2shape}& $48.72$ & $44.80$ & $70.13$ & $62.5$ & $41.1$ \\
    BRepMAE~\cite{brepmae}& $74.94{\pm}0.45$ & $80.09{\pm}0.34$ & $76.52{\pm}0.71$ & $66.5{\pm}0.1$ & $46.6{\pm}0.1$ \\
    \midrule
    Same-corruption reconstruction & $77.63{\pm}1.02$ & $80.43{\pm}0.63$ & $79.17{\pm}0.38$ & $67.8{\pm}0.0$ & $48.6{\pm}0.0$ \\
    Graph-JEPA objective & $78.68{\pm}3.15$ & $80.07{\pm}4.86$ & $75.00{\pm}1.71$ & $63.9{\pm}0.7$ & $43.3{\pm}1.0$ \\
    \midrule
    \method (ours)
    & \cellcolor{oursgray}{\boldmath$81.29{\pm}0.39$}
    & \cellcolor{oursgray}{\boldmath$89.61{\pm}0.32$}
    & \cellcolor{oursgray}{\boldmath$80.08{\pm}0.49$}
    & \cellcolor{oursgray}{\boldmath$69.1{\pm}0.1$}
    & \cellcolor{oursgray}{\boldmath$50.4{\pm}0.2$} \\
    \bottomrule
  \end{tabularx}
\end{table*}

\paragraph{Pretraining.}
All methods are pretrained on the same 100{,}000 unlabeled ABC solids~\cite{abc}. We remove near-duplicates and solids that are geometrically identical to any evaluation solid. Each \method run takes $362.5$k optimization steps with batch size $16$ and AdamW~\cite{loshchilov2019adamw} (learning rate $10^{-3}$, weight decay $0.05$), with a $5\%$ warm-up, cosine learning-rate decay~\cite{loshchilov2017sgdr}, and gradient clipping at $1.0$. We use a two-layer transformer with hidden dimension $384$ and six attention heads for the predictor. The projector is a two-layer MLP with $4\times$ expansion.

\paragraph{Baselines.}
We compare \method against state-of-the-art self-supervised learning approaches using their official implementations, architectures and training schedules on our pretraining dataset. More specifically, we compare: (i)~BRepMAE~\cite{brepmae}, (ii)~Brep2Shape~\cite{brep2shape} and (iii)~contrastive UV-Net~\cite{uvnet}. We also evaluate Graph-JEPA~\cite{graphjepa}, whose loss becomes non-finite on B-Rep graphs across all seeds, despite the row normalization provided by its implementation. We therefore use its last checkpoints with finite loss (13 of 30 epochs, supplementary material).

We conduct additional experiments to isolate the pretraining objective. To control for the masking strategy, we evaluate reconstruction using the same encoder and masks as \method (\emph{Same-corruption reconstruction} in Table~\ref{tab:comparison}). The \emph{Graph-JEPA objective}~\cite{graphjepa} is adapted to our encoder using its official patch construction and it follows the original optimization steps (METIS~\cite{karypis1998metis} randomization and hyperbolic targets). Finally, \emph{random initialization} provides an untrained encoder reference. We use three pretraining seeds for all methods, except Brep2Shape and UV-Net, which use one seed each. We report mean $\pm$ standard deviation across seeds where applicable.

\paragraph{Tasks and protocol.}
\Cref{fig:benchmarks} presents the four downstream tasks. We evaluate eight-class Fusion~360 Gallery face segmentation~\cite{brepnet}, 25-class MFInstSeg machining-feature segmentation~\cite{mfinstseg} and ten-class TMCAD part classification~\cite{zou2025brt}, with $5{,}366$, $9{,}375$, and $1{,}626$ test solids, respectively. Fusion~360 uses its official split. We evaluate frozen encoders using linear probes~\cite{chen2020simclr,he2020moco}: 256-dimensional embeddings are standardized and classified using $\ell_2$-regularized logistic regression, with regularization strength selected on validation. TMCAD uses mean-pooled face embeddings, while MFInstSeg uses fixed subsets of $8{,}000$ training and $2{,}000$ validation solids. Macro-F1 is the primary metric.

For BrepEDIT~\cite{liu2026brepler}, a cross-encoder head operating on frozen embeddings ranks candidate edits of the same source part given the source part and a geometric instruction. We evaluate on our leakage-free split of $2{,}000$ test parts. All downstream heads use the same training protocol across encoders, with learning rates selected on validation. We report mean reciprocal rank (MRR) and Recall@1 (R@1). Both metrics are multiplied by $100$. Full evaluation details are provided in the supplementary material.

\begin{figure*}[t]
  \centering
  \includegraphics[width=\textwidth]{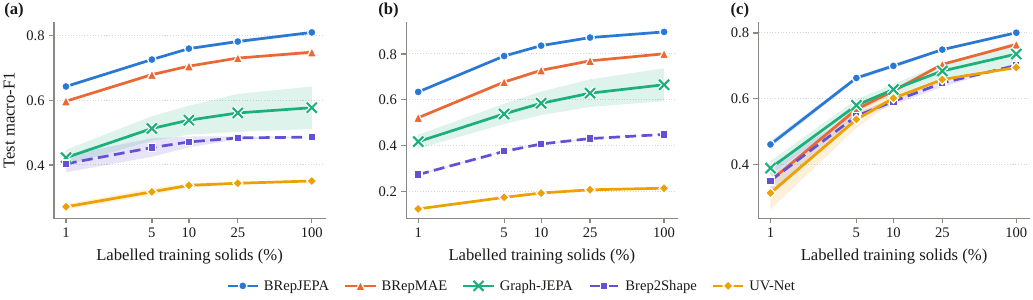}
  \caption{\textbf{Label efficiency.} Test macro-F1 of frozen linear probes trained on $1$--$100\%$ of the training solids (logarithmic x-axis) for (a)~Fusion~360, (b)~MFInstSeg, and (c)~TMCAD. Training subsets are nested and shared across encoders. Lines show performance averaged over pretraining seeds and three subset draws. Shaded bands indicate one standard deviation across seeds, or across subset draws for single-seed methods (Brep2Shape and UV-Net). Graph-JEPA uses its official implementation.}
  \label{fig:labeleff}
  \vspace{-4pt}
\end{figure*}

\subsection{Comparison with Pretraining Objectives}
\label{sec:comparison}
\begin{table*}[t]
  \centering
  \caption{\textbf{Fine-tuning with few labels.} Full fine-tuning of the encoder and a segmentation head on nested random subsets of $1$, $5$, and $10\%$ of the training solids and on all of them (test macro-F1 in \%). One recipe serves every method. MFInstSeg $100\%$ uses the probe protocol's $8{,}000$ training solids.}
  \label{tab:finetune}
  \vspace{-4pt}
  \footnotesize
  \setlength{\tabcolsep}{2pt}
  \setlength{\aboverulesep}{0.3ex}\setlength{\belowrulesep}{0.5ex}\renewcommand{\arraystretch}{1.1}
  \begin{tabular}{@{}lcccccccc@{}}
    \toprule
    & \multicolumn{4}{c}{Fusion~360} & \multicolumn{4}{c}{MFInstSeg} \\
    \cmidrule(lr){2-5}\cmidrule(l){6-9}
    Method & $1\%$ & $5\%$ & $10\%$ & $100\%$ & $1\%$ & $5\%$ & $10\%$ & $100\%$ \\
    \midrule
    Random initialization & $69.57{\pm}0.15$ & $80.68{\pm}0.43$ & $83.83{\pm}0.14$ & $90.93{\pm}0.59$ & $78.77{\pm}0.95$ & $92.02{\pm}0.05$ & $94.19{\pm}0.11$ & $98.22{\pm}0.10$ \\
    \midrule
    UV-Net~\cite{uvnet} & $38.05{\pm}1.64$ & $47.54{\pm}1.55$ & $51.46{\pm}0.98$ & $62.91$ & $35.49{\pm}0.20$ & $52.69{\pm}1.37$ & $58.82{\pm}0.97$ & $85.40$ \\
    Graph-JEPA~\cite{graphjepa} & $65.16{\pm}1.76$ & $75.96{\pm}0.73$ & $78.53{\pm}1.40$ & $88.59{\pm}1.12$ & $62.73{\pm}7.21$ & $87.85{\pm}1.19$ & $91.75{\pm}0.71$ & $97.64{\pm}0.10$ \\
    Brep2Shape~\cite{brep2shape} & $59.58{\pm}1.69$ & $72.99{\pm}0.64$ & $77.03{\pm}0.91$ & $88.00$ & $49.93{\pm}2.26$ & $85.27{\pm}1.63$ & $90.94{\pm}0.38$ & $97.38$ \\
    BRepMAE~\cite{brepmae} & $67.27{\pm}0.45$ & $77.49{\pm}0.20$ & $80.37{\pm}0.44$ & $87.79{\pm}0.27$ & {$82.05{\pm}0.67$} & {\boldmath$94.71{\pm}0.16$} & {\boldmath$96.16{\pm}0.10$} & {\boldmath$98.73{\pm}0.14$} \\
    \midrule
    Same-corruption recon. & $70.21{\pm}0.35$ & $82.20{\pm}0.45$ & $85.33{\pm}0.36$ & $91.11{\pm}0.39$ & $78.49{\pm}0.21$ & $91.98{\pm}0.17$ & $94.36{\pm}0.18$ & $98.19{\pm}0.21$ \\
    Graph-JEPA objective & $69.97{\pm}0.31$ & $80.54{\pm}0.67$ & $83.65{\pm}0.20$ & $90.99{\pm}0.09$ & $79.75{\pm}0.29$ & $91.95{\pm}0.05$ & $94.12{\pm}0.08$ & $98.26{\pm}0.01$ \\
    \midrule
    \method (ours)
    & \cellcolor{oursgray}{\boldmath$71.73{\pm}0.33$}
    & \cellcolor{oursgray}{\boldmath$82.93{\pm}0.12$}
    & \cellcolor{oursgray}{\boldmath$85.67{\pm}0.12$}
    & \cellcolor{oursgray}{\boldmath$91.47{\pm}0.23$}
    & \cellcolor{oursgray}{\boldmath$82.40{\pm}0.38$}
    & \cellcolor{oursgray}{$93.17{\pm}0.13$}
    & \cellcolor{oursgray}{$95.15{\pm}0.09$}
    & \cellcolor{oursgray}{$98.36{\pm}0.30$} \\
    \bottomrule
  \end{tabular}
\end{table*}

\Cref{tab:comparison} compares all methods using frozen encoders. BRepMAE is the strongest published model. \method leads it by $+6.3$ macro-F1 points on Fusion~360, $+9.5$ on MFInstSeg, and $+3.6$ on TMCAD. Brep2Shape, UV-Net, and the official Graph-JEPA are far behind on the two face-level benchmarks and $6$--$12$ points behind on TMCAD. The second group isolates the objective. On MFInstSeg, \method outperforms same-corruption reconstruction by $+9.19$ macro-F1 points. This baseline uses exactly the masks of \method, demonstrating that the choice of pretraining objective matters. On Fusion~360, \method improves over same-corruption reconstruction by $+3.66$ and on TMCAD by $+0.91$. 

\method outperforms all methods on edit retrieval as well. On BrepEDIT, \method leads same-corruption reconstruction by $+1.3$ MRR points and BRepMAE by $+2.7$. The Graph-JEPA objective falls below same-corruption reconstruction on MFInstSeg and TMCAD and varies most across seeds.

\subsection{Label Efficiency}
\label{sec:labeleff}

Self-supervised pretraining matters most when labels are scarce. Results of frozen encoder probes trained on nested subsets of $1$--$100\%$ of the training solids are presented in \Cref{fig:labeleff}. \method outperforms every published method at every label fraction. It leads BRepMAE by $4.5$--$6.1$ points on Fusion~360 and $9.5$--$11.3$ on MFInstSeg. Against same-corruption reconstruction on our encoder, its margin grows as labels become scarcer. With $1\%$ of the labels, it reaches $+18.8$ points on MFInstSeg ($80$ training solids), $+7.0$ on TMCAD ($76$ solids) and $+5.6$ on Fusion~360 ($212$ solids).

\begin{table*}[t]
  \centering
  \caption{\textbf{Per-class results.} Test F1 of frozen linear probes for every class of Fusion~360 and MFInstSeg, mean over three pretraining seeds.}
  \label{tab:perclass}
  \vspace{-4pt}
  \footnotesize
  \setlength{\tabcolsep}{3pt}
  \setlength{\aboverulesep}{0pt}\setlength{\belowrulesep}{0pt}\renewcommand{\arraystretch}{1.12}
  \begin{tabularx}{0.95\textwidth}{@{}l*{2}{>{\centering\arraybackslash}X}@{\hspace{12pt}}l*{2}{>{\centering\arraybackslash}X}@{\hspace{12pt}}l*{2}{>{\centering\arraybackslash}X}@{}}
    \toprule
    Fusion~360 & BRepMAE & \method & MFInstSeg & BRepMAE & \method & MFInstSeg & BRepMAE & \method \\
    \midrule
    ExtrudeSide & $88.6$ & \cellcolor{oursgray}{\boldmath$90.6$} & Chamfer & $91.6$ & \cellcolor{oursgray}{\boldmath$92.4$} & Tri. pocket & $87.2$ & \cellcolor{oursgray}{\boldmath$98.2$} \\
    ExtrudeEnd & $88.1$ & \cellcolor{oursgray}{\boldmath$90.0$} & Through hole & $95.6$ & \cellcolor{oursgray}{\boldmath$97.4$} & Rect. pocket & $87.4$ & \cellcolor{oursgray}{\boldmath$95.5$} \\
    CutSide & $62.6$ & \cellcolor{oursgray}{\boldmath$67.2$} & Tri. passage & $65.2$ & \cellcolor{oursgray}{\boldmath$96.9$} & 6-sided pocket & $92.5$ & \cellcolor{oursgray}{\boldmath$98.6$} \\
    CutEnd & $67.0$ & \cellcolor{oursgray}{\boldmath$77.5$} & Rect. passage & $79.1$ & \cellcolor{oursgray}{\boldmath$89.7$} & Circ. end pocket & $94.5$ & \cellcolor{oursgray}{\boldmath$97.6$} \\
    Fillet & $89.5$ & \cellcolor{oursgray}{\boldmath$97.3$} & 6-sided passage & $84.0$ & \cellcolor{oursgray}{\boldmath$98.0$} & Rect. blind slot & $37.2$ & \cellcolor{oursgray}{\boldmath$68.8$} \\
    Chamfer & $86.1$ & \cellcolor{oursgray}{\boldmath$92.4$} & Tri. through slot & $59.8$ & \cellcolor{oursgray}{\boldmath$80.1$} & Vert. circ. end blind slot & $74.6$ & \cellcolor{oursgray}{\boldmath$88.7$} \\
    RevolveSide & $75.5$ & \cellcolor{oursgray}{\boldmath$80.3$} & Rect. through slot & $37.4$ & \cellcolor{oursgray}{\boldmath$53.3$} & Horiz. circ. end blind slot & $86.1$ & \cellcolor{oursgray}{\boldmath$95.0$} \\
    RevolveEnd & $42.0$ & \cellcolor{oursgray}{\boldmath$55.2$} & Circ. through slot & $90.3$ & \cellcolor{oursgray}{\boldmath$92.5$} & Tri. blind step & $81.9$ & \cellcolor{oursgray}{\boldmath$88.6$} \\
     &  &  & Rect. through step & $68.6$ & \cellcolor{oursgray}{\boldmath$74.8$} & Circ. blind step & $95.7$ & \cellcolor{oursgray}{\boldmath$98.1$} \\
     &  &  & 2-sided through step & $75.1$ & \cellcolor{oursgray}{\boldmath$86.3$} & Rect. blind step & $74.0$ & \cellcolor{oursgray}{\boldmath$83.5$} \\
     &  &  & Slanted through step & $64.7$ & \cellcolor{oursgray}{\boldmath$77.4$} & Round & $97.8$ & \cellcolor{oursgray}{\boldmath$98.6$} \\
     &  &  & O-ring & $93.7$ & \cellcolor{oursgray}{\boldmath$96.5$} & Stock & $99.0$ & \cellcolor{oursgray}{\boldmath$99.8$} \\
     &  &  & Blind hole & $89.7$ & \cellcolor{oursgray}{\boldmath$94.3$} &  &  &  \\
    \bottomrule
  \end{tabularx}
\end{table*}

\subsection{Ablation Study}
\label{sec:ablations}

\begin{figure*}[t]
  \centering
  \includegraphics[width=\textwidth]{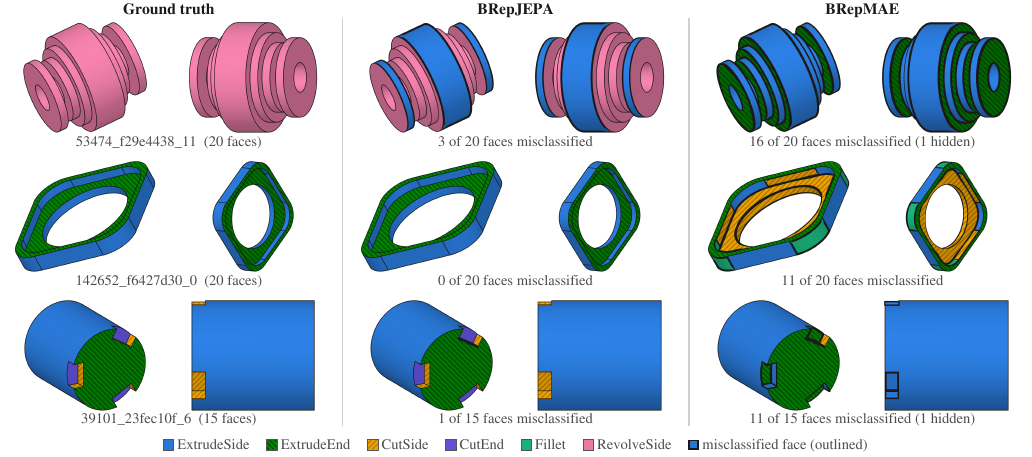}
  \caption{\textbf{Qualitative results.} Fusion~360 face segmentation on official test parts: ground truth and frozen linear-probe predictions (seed $42$) of \method and BRepMAE, two views per part, misclassified faces outlined.}
  \label{fig:qualitative}
  \vspace{-4pt}
\end{figure*}

\begin{table}[t]
  \centering
  \caption{\textbf{Ablations.} Each row changes one component of \method. Test macro-F1 (\%) of frozen linear probes.}
  \label{tab:ablation}
  \vspace{-4pt}
  \footnotesize
  \setlength{\tabcolsep}{1.5pt}
  \setlength{\aboverulesep}{0pt}\setlength{\belowrulesep}{0pt}\renewcommand{\arraystretch}{1.12}
  \begin{tabularx}{\columnwidth}{@{}l*{3}{>{\centering\arraybackslash}X}@{}}
    \toprule
    Variant & Fusion~360 & MFInstSeg & TMCAD \\
    \midrule
    Final teacher layer only & $77.66{\pm}0.42$ & $84.11{\pm}0.30$ & $79.00{\pm}0.48$ \\
    Half as many targets & $79.72{\pm}0.30$ & $89.55{\pm}0.06$ & $79.17{\pm}0.11$ \\
    VICReg instead of SIGReg & $79.37{\pm}0.69$ & $89.76{\pm}0.29$ & $80.00{\pm}0.66$ \\
    No regularizer & $79.72{\pm}0.58$ & \boldmath$89.81{\pm}0.57$ & $79.79{\pm}0.55$ \\
    \midrule
    \method
    & \cellcolor{oursgray}{\boldmath$81.29{\pm}0.39$}
    & \cellcolor{oursgray}{$89.61{\pm}0.32$}
    & \cellcolor{oursgray}{\boldmath$80.08{\pm}0.49$} \\
    \bottomrule
  \end{tabularx}
\end{table}

In \Cref{tab:ablation} we report experiments in which one component of \method is removed or replaced at a time. Additional variants are provided in the supplementary material.

\paragraph{Teacher depth.}
Across three seeds, using the normalized final-layer representations as targets reduces performance by $5.5$ points on MFInstSeg and $3.6$ on Fusion~360, indicating that the benefit of multi-depth targets is not simply due to normalization. In single-seed experiments, subsets combining early and late layers perform similarly to the full average. Excluding the early layers (\{2,3,4\}) leads to the largest drop. This is consistent with early representations preserving local geometric information that may be lost in deeper, more context-dependent representations.
\paragraph{Target design.}
Random local targets, which match the size, anchors, and per-graph masked-entity budget of the topology-defined targets, perform worse across three seeds by $1.47$ points on Fusion~360, $0.66$ on MFInstSeg and $0.63$ on TMCAD. The topology-defined targets thus provide a small but consistent gain, while most of the benefit over reconstruction comes from predicting entity-level latents.
\paragraph{Regularizer.}
Replacing SIGReg by prediction-level variance--covariance regularization~\cite{vicreg} lowers Fusion~360 by $1.92$ points over three seeds ($[0.42, 3.43]$) and leaves MFInstSeg and TMCAD within noise. SIGReg acts on the online embeddings rather than on the predictions and regularizes the full embedding distribution instead of its first two moments. Without any regularizer, training remains stable and does not collapse completely, but the online embeddings lose about $70\%$ of their effective rank on every seed and Fusion~360 drops by $1.57$ points ($[0.10, 3.09]$), with MFInstSeg and TMCAD within noise (supplementary material).

\subsection{Fine-Tuning}
\label{sec:finetune}
When the full encoder is fine-tuned end to end using all available labels, differences between pretraining objectives shrink (\cref{tab:finetune}). On MFInstSeg, performance saturates near $98\%$ for every initialization of our encoder. On Fusion~360, \method reaches $91.47$, exceeding same-corruption reconstruction by $+0.36$ points. These results suggest that the benefits of pretraining are more significant with frozen representations or limited labeled data.

Under limited supervision, the fine-tuning gaps widen again. \method outperforms every baseline using our encoder by $0.8$--$1.2$ points on MFInstSeg with $5\%$ or $10\%$ of the training solids. On Fusion~360, it exceeds the Graph-JEPA objective by $1.8$--$2.4$ points at $1$--$10\%$ supervision.

Among published methods, BRepMAE achieves the strongest fine-tuning results on MFInstSeg, outperforming \method by $1.0$--$1.5$ points at $5$--$10\%$ supervision. At $1\%$, however, \method leads BRepMAE by $0.35$ points. The BRepMAE architecture trained from random initialization already reaches $80.50$, $94.14$, and $95.81$ at these label fractions (see supplementary material). On Fusion~360, \method outperforms BRepMAE by $3.7$--$5.4$ points at every label fraction.

\subsection{Per-Class Analysis}
\label{sec:perclass}
\Cref{tab:perclass} reports per-class F1 scores. On MFInstSeg, the improvements are not confined to a few categories: \method outperforms BRepMAE across all $25$ classes, with gains ranging from less than one point on Stock, Chamfer, and Round to more than $30$ points on triangular passages and rectangular blind slots. On Fusion~360, \method improves performance across all eight classes, with the largest gains on RevolveEnd and CutEnd (more than $10$ points). \Cref{fig:qualitative} illustrates three Fusion~360 test parts. In these examples, BRepMAE frequently misclassifies revolved faces as extruded faces and confuses extruded faces with cut faces.
\section{Discussion and Conclusion}
\label{sec:conclusion}
We introduced \method, the first joint-embedding predictive pretraining method for B-Reps, which learns representations by predicting the latents of topology-defined targets from multi-depth teacher states. Pretrained on the same corpus, its frozen representations outperform published B-Rep pretraining methods. With a shared encoder, \method also outperforms same-corruption reconstruction and the Graph-JEPA objective on face segmentation and part classification, with the largest gains under limited supervision, and leads on edit-outcome retrieval. Our ablations show that multi-depth targets account for much of the improvement, while topology-defined targets provide a smaller but consistent gain over random targets of matched size.

These findings are limited to single-solid ABC models with $4$--$200$ faces and one encoder architecture. The use of absolute coordinates also retains the orientation and face-partition sensitivities of B-Rep networks~\cite{ballegeer2026fovnet,jiang2026canonical}. Comparisons are based on three pretraining seeds, and differences between pretraining objectives largely vanish under full fine-tuning with all labels.

\section*{LLM Usage Statement}
In this work, Large Language Models (LLMs) were utilized to support both software development and manuscript preparation. Specifically, Claude 5.5 was used as a general purpose tool to assist with boilerplate code and GPT-6.1-Sol was employed to assist with editing prose for clarity, flow, and grammatical correctness. No LLM or AI tool was used to develop the research ideas or hypotheses, formulate the scientific claims, or draw the conclusions. Neither the full manuscript nor substantial parts of it were submitted to an LLM provider for reviewer style feedback. The authors take full responsibility for all submitted content, including material prepared with AI assistance.

{
    \small
    \bibliographystyle{ieeenat_fullname}
    \bibliography{references_all}
}

\onecolumn
\raggedbottom

\setcounter{section}{0}
\setcounter{subsection}{0}
\setcounter{figure}{0}
\setcounter{table}{0}
\setcounter{equation}{0}

\renewcommand{\thesection}{S\arabic{section}}
\renewcommand{\thefigure}{S\arabic{figure}}
\renewcommand{\thetable}{S\arabic{table}}
\renewcommand{\theequation}{S\arabic{equation}}

\renewcommand{\theHsection}{supp.\arabic{section}}
\renewcommand{\theHfigure}{supp.\arabic{figure}}
\renewcommand{\theHtable}{supp.\arabic{table}}
\renewcommand{\theHequation}{supp.\arabic{equation}}

\renewcommand{\topfraction}{0.92}
\renewcommand{\bottomfraction}{0.8}
\renewcommand{\textfraction}{0.06}
\renewcommand{\floatpagefraction}{0.75}

\setcounter{topnumber}{4}
\setcounter{bottomnumber}{3}
\setcounter{totalnumber}{5}

\setlength{\aboverulesep}{0.35ex}
\setlength{\belowrulesep}{0.35ex}
\setlength{\textfloatsep}{13pt plus 2pt minus 2pt}
\setlength{\intextsep}{12pt plus 2pt minus 2pt}
\setlength{\hfuzz}{0.5pt}
\setlength{\vfuzz}{0.5pt}

\begin{center}
    {\LARGE\bfseries
    BRepJEPA: Predictive Self-Supervised Learning\\
    on Heterogeneous B-Rep Graphs\par}
    \vspace{8pt}
    {\Large\bfseries Supplementary Material\par}
\end{center}

\vspace{10pt}

\noindent
\method uses the topology of a boundary representation to define what a student encoder should predict. In this supplement, we specify the construction of those targets and the downstream evaluations. We also examine how the learned representations behave when the supervision, target design or evaluation changes. \Cref{sec:s-method,sec:s-protocols} give the pretraining and evaluation protocols. \Cref{sec:s-results,sec:s-ablations} report extended transfer results and ablations.

\section{Pretraining on B-Reps}
\label{sec:s-method}

\subsection{Geometry and incidence}
We follow the graph modeling of TopoGNN~\cite{topognn}. Each solid is a graph with face, edge and vertex nodes. The eight directed relations are face adjacency, loop succession, face-to-edge and edge-to-face incidence split by coedge orientation, and edge-to-vertex and vertex-to-edge incidence. A coedge is one oriented use of an edge by a face. A seam can contribute two such uses to the same face. For the loop and face--edge relations we record up to eight coedge slots. Face adjacency and edge--vertex incidence are stored without this cap.

Face features consist of a $16{\times}16$ UV grid (three position channels, three normal channels and one trimming mask), a $10{\times}10$ net of weighted NURBS control points and 17 handcrafted descriptors. The face encoder combines a four-layer CNN on the grid, a point MLP with masked  mean pooling on the control net and a descriptor MLP. Their outputs are then fused with a two-layer MLP and LayerNorm. Edges use a one-dimensional CNN on up to 20 weighted curve control points and an MLP on eight descriptors. The resulting 256-dimensional entity tokens pass through four pre-normalized residual GATv2 layers~\cite{brody2022gatv2}. Each layer has eight 32-dimensional heads per relation and averages the relation outputs. Dropout and attention dropout are both $0.1$. The encoder has $4.99$\,M parameters. Retaining all relation types, including empty ones, makes the frozen readout independent of the other solids in its batch.

\subsection{Target sampling and the online view}
\label{sec:s-sampling}

A solid contributes up to two loops, four coedges, one region boundary and three faces, with targets redrawn every step. In \Cref{tab:s-target-design}, the sampling rule and the features hidden by each draw are presented. When a solid has too few eligible candidates, it contributes fewer targets. Over 1,008 sampled ABC solids, the shortfall was $1.7\%$ of loop requests, $0\%$ of coedge requests, $10.5\%$ of region requests and $0.5\%$ of face requests. The mean was $9.9$ targets per solid, with $21.7\%$ of faces and $22.0\%$ of edges hidden in a batch of 16 solids.

\begin{table}[htbp]
  \centering
  \caption{\textbf{Targets drawn from each solid.} The teacher encodes the complete graph. The last column gives the features each draw hides from the student before the masks are combined.}
  \label{tab:s-target-design}
  \small
  \begin{tabularx}{\linewidth}{@{}lX>{\raggedright\arraybackslash}p{0.27\linewidth}@{}}
    \toprule
    Family & Sampling and teacher target & Hidden features \\
    \midrule
    Loop
      & Draw a recorded loop uniformly from loops of at least three edge entries. Skip loops sharing an edge with an earlier loop draw. Average the teacher latents of the loop edges.
      & Loop edges, their end vertices and the incident face. \\
    Coedge
      & Draw an oriented face--edge incidence uniformly, using each edge at most once. Apply a fixed orientation-conditioned map to the teacher latents of its edge and face.
      & The edge, its end vertices, and the sampled incident face. \\
    Region boundary
      & Grow a region from a uniform anchor face by a uniform radius of one to three face-adjacency hops. Accept a proper region of 3--25 faces with at least three recorded boundary edges. Average the teacher latents of those edges.
      & Recorded boundary edges and their end vertices. \\
    Face
      & Draw uniformly without replacement among faces with at least two adjacent faces. Use the face's teacher latent.
      & The sampled face. \\
    \bottomrule
  \end{tabularx}
\end{table}

The student makes one pass through the solid with the union of the masks. It receives a learned token of the corresponding entity type in place of each hidden initial embedding. Face--edge, edge--vertex, and face-adjacency relations remain available. Loop-succession links touching a masked edge are removed, and masking the end vertices with the edge removes direct access to endpoint coordinates. The region draw leaves its interior face features available to the student unless another draw masks them. When requesting $n$ regions, the sampler tries at most $8n$ draws and discards repeated boundary sets. 
\FloatBarrier

\subsection{Multi-depth teacher and prediction}
The teacher averages parameter-free LayerNorm outputs from its input encoder and all four message-passing layers before projection. For a face or edge $u$, its projected latent is
\begin{equation}
  \bar z_u = p_{\bar\theta}\!\left(\frac{1}{5}\sum_{\ell=0}^{4}
  \operatorname{LN}\!\left(h^{(\ell)}_{u,\bar\theta}\right)\right),
  \label{eq:s-teacher}
\end{equation}
where $\bar\theta$ denotes the EMA parameters. The student projects its final layer. The projectors are two-layer MLPs ($256\to1024\to256$, with LayerNorm and GELU). A coedge target uses $g([\bar z_e,\bar z_f,\eta_o])$, where $\eta_o$ is the orientation embedding and $g$ is a $768\to512\to256$ MLP with LayerNorm. Both $\eta_o$ and $g$ are fixed at random initialization. For loop and boundary representations, a preliminary single-seed version with frozen attention pooling improved macro-F1 by $0.1$ points relative to mean pooling. The reported model uses mean pooling.

Queries contain a learned \textsc{cls} token, the projected student embedding of the anchor face, target-family, orientation embeddings and the projected student token of each target entity with its entity-type embedding added. These entity tokens are formed after message passing on the masked graph. They identify the masked entities in each query, including when two targets share their anchor, family, and orientation. The predictor retains at most 64 evenly spaced entity tokens per target and processes the full sequence with two pre-normalized transformer layers of width 384, six heads, feed-forward width 1,536, GELU, and dropout $0.1$. A LayerNorm and a two-layer MLP map the updated \textsc{cls} token to one 256-dimensional vector per target.

The prediction loss is Smooth-L1 against teacher targets with stopped gradients. It is averaged first within each family and then across the families present. For regularization, we apply SIGReg~\cite{balestriero2025lejepa} separately to all projected online face and edge embeddings, including those of masked entities. At each step, SIGReg draws 1,024 Gaussian directions, normalizes them, and shares them across the two entity types. We use the Epps--Pulley statistic~\cite{epps1983normality} along each direction to compare the empirical characteristic function with that of a standard Gaussian. Trapezoidal integration uses 17 equispaced points on $[0,3]$ and symmetry. For $N$ entities of a given type across $B$ solids, the statistic is averaged over $\lceil N/B\rceil$ groups, each containing one entity from every contributing solid. We randomly order the entities within each solid and take the first $K$, cycling through the list if it contains fewer than $K$. The total loss is $0.95$ times the prediction loss plus $0.05$ times the mean of the two SIGReg terms. The teacher stays in evaluation mode and its encoder and projector are updated by EMA after every step. Downstream evaluation uses the final 256-dimensional state of the teacher encoder.

\subsection{Pretraining dataset, schedule and cost}
For pretraining we use 100,000 single-solid ABC models~\cite{abc} with 4--200 faces after graph extraction. We removed 6,736 near-duplicate ABC candidates using a geometric-signature audit, which found 709 geometric matches to Fusion~360 and 10 to TMCAD. None of the 5,366 Fusion~360 test solids has a geometric twin in the retained dataset. The BrepEDIT split adds a separate check for overlap with the ABC pretraining selection (\cref{sec:s-edit}).

We use AdamW with $\beta=(0.9,0.95)$, weight decay $0.05$, a learning rate of $10^{-3}$, three warm-up epochs, cosine decay to zero and global gradient clipping at $1.0$. The batch holds 16 solids. There are 58 passes through ABC (362,500 updates), with EMA decay increasing from $0.996$ to $1$ on a cosine schedule. The trainable projector and predictor add $4.43$\,M parameters for pretraining. Each step evaluates both the masked student and the complete-graph teacher. Across the three seeds, a run took $18.1$ hours on average with $3.6$--$3.8$\,GB peak allocated memory on a shared RTX~4090. The same-corruption reconstruction baseline ran for the same number of updates and took $9.8$--$12.6$ hours. Shared GPU load limits interpretation of these timings as a controlled comparison of cost.

\section{Evaluation and comparison protocols}
\label{sec:s-protocols}

\subsection{Probes, label budgets and fine-tuning}
\label{sec:s-probes}
For the shared encoder, a frozen probe reads the final 256-dimensional embedding of each face. On part classification (TMCAD) it averages the face embeddings of each solid before fitting the probe. We standardize with training statistics and fit multinomial logistic regression with L-BFGS for at most 1,000 iterations. We choose the $\ell_2$ regularization parameter $C$ from $\{0.01,0.1,1,10\}$ by validation macro-F1, then score the chosen probe on test. Published architectures use the readouts in \cref{sec:s-published,sec:s-graphjepa} and the same probe protocol.


For Fusion~360 we use the official dataset test split. We fit the probe on the first $\operatorname{round}(fN)$ solids of a seeded permutation of the training list at each reduced label fraction ($1$, $5$, $10$, and $25\%$). This gives us nested subsets that are shared across the encoders, with three draws for the published models comparison and ten for the shared encoder comparison. The full validation split selects $C$ at every fraction.

For fine-tuning, we update the encoder with the attached face-segmentation MLP ($256\to256\to128\to C$ for $C$ classes, dropout $0.1$) using cross-entropy. The official Graph-JEPA uses 1,024 dimensional face embeddings and the other compared models use their published embedding dimensions. All methods use AdamW, weight decay $0.05$, gradient clipping at $1.0$ and one warm-up epoch followed by cosine decay. The learning rate is selected from $\{10^{-4},5{\cdot}10^{-4}\}$ on validation with seed 42 and reused across pretraining seeds. We allow 40 epochs on Fusion~360 and 50 on MFInstSeg, with patience 10 and selection by best validation macro-F1. MFInstSeg uses the same 8,000-solid training set as the probes. For the $1$, $5$, and $10\%$ fine-tuning experiments, we use three shared subset draws, choose the learning rate on the first draw of each fraction, and validate every $\max(\text{one epoch},R)$ updates with $R=125$ on Fusion~360 and $R=100$ on MFInstSeg. This gives a budget of at least 5,000 updates before early stopping.

\subsection{Edit-outcome retrieval}
\label{sec:s-edit}
For the edit-outcome retrieval task, we use the BrepEDIT-240K~\cite{liu2026brepler} dataset. It contains language instructions with single-face deletions and their inverses. Our evaluation task uses the deletion direction. Given a source solid and a geometric instruction, the retrieval head ranks candidate edit outcomes. Training, validation and test splits contain $34{,}212/500/2{,}000$ source parts and $72{,}928/2{,}220/8{,}937$ edits. The test split includes 128 parts with a non-identical near-duplicate in training.


All encoders use the same retrieval head (four blocks, width 256, $3.3$\,M parameters). Its sequence contains a classification token, a frozen instruction embedding from \texttt{bge-small-en-v1.5} and the source and candidate face tokens built from standardized frozen embeddings. Training samples instructions across all four styles and uses a softmax over the correct outcome and eight edits from other parts matched by face count. The learning rate ($10^{-4}$ or $3{\cdot}10^{-4}$) and epoch (up to 20) are selected on validation with three head seeds. Test metrics use three geometric instructions per edit. Scoring excludes siblings that match the target's face, edge and vertex counts and agree in area and volume to the matching tolerance. We report MRR and Recall@1 multiplied by 100, with same-part chance scores of $47.7$ and $22.4$, respectively. The detailed, functional, and concise instructions are also scored in \cref{tab:s-secondary}.


\subsection{Statistics}
For comparisons with shared-encoder models, we use three pretraining seeds. Means and sample standard deviations are reported across pretraining seeds unless stated otherwise. The $95\%$ percentile bootstrap confidence intervals for shared-encoder models are computed by resampling seed pairs and subsequently resampling test solids, with macro-F1 recomputed from per-solid confusion counts. For retrieval, we resample source parts. For reduced-label experiments, subset draws are resampled between seed pairs and test units. We use 2,000 bootstrap replicates for frozen probes and 10,000 for fine-tuning and retrieval. With three seeds, there are ten distinct multisets of resampled seed pairs.

\subsection{Published B-Rep encoders}
\label{sec:s-published}
BRepMAE~\cite{brepmae}, Brep2Shape~\cite{brep2shape} and UV-Net~\cite{uvnet} use their released architectures and pretraining objectives on ABC. We keep each model's training schedule and feature representation. \Cref{tab:s3-budgets} reports training budgets.
\begin{table*}[htbp]
  \centering
  \caption{\textbf{Pretraining budgets.} Parameters count the modules used for the evaluated embedding and all trainable parameters, respectively. Wall-clock times are from shared GPUs. BRepMAE compute is for the complete seed-42 schedule.}
  \label{tab:s3-budgets}
  \footnotesize
  \setlength{\tabcolsep}{5pt}
  \begin{tabular}{@{}llcccccccc@{}}
  \toprule
  & & Training & & & & \multicolumn{2}{c}{Parameters (M)} & & \\
  \cmidrule(lr){7-8}
  Method & Code & Batch & Passes & Updates & Encoder & Total & GPU & Hours \\
  \midrule
  \method & ours & 16 & 58 & 362.5k & 4.99 & 9.42 & RTX 4090 & 18.1 \\
  BRepMAE & official & 256 & 150 & 53k & 3.78 & 8.03 & RTX PRO 6000 & 19.1 \\
  Brep2Shape & official & $4{\times}32$ & 100 & 61k & 4.13 & 4.19 & RTX 4090 & 31.6 \\
  UV-Net & official & 256 & 350 & 129k & 1.13 & 1.17 & Tesla P100 & 33.6 \\
  \bottomrule
  \end{tabular}
\end{table*}

For BRepMAE we use the official extractor, feature normalization fitted on ABC and pretraining schedule (batch 256, mask rate $0.8$, learning rate $10^{-4}$, up to 150 epochs, patience 25). The per-face readout is the encoder output with the virtual node dropped. For TMCAD, validation selects between the virtual node and the mean face embedding. 

For Brep2Shape we use the official pretraining code and the BRT tokenizer~\cite{zou2025brt} for STEP preprocessing. The face graph, line graph, and targets reproduce the released checkpoint's reconstruction loss. The training configuration comes from that checkpoint (six graph, three surface, and three edge layers, width 128, four heads, 100 epochs, AdamW at $8{\cdot}10^{-4}$ with cosine decay and clipping at $1.0$). Four accumulated batches of 32 solids give 128 solids per update. The last Dual Transformer block supplies the face readout, which is averaged for TMCAD. 

Finally, for UV-Net we use the official self-supervised branch and graph builder ($10{\times}10$ face grids and 10 curve samples). The contrastive recipe uses width 128, projection to 64 dimensions, temperature $0.1$, the official view transformations, and 350 epochs with batch size 256 and Adam with weight decay $10^{-4}$. We use the minimum-validation-loss checkpoint. Validation selects the concatenation of the node and shape embeddings for segmentation and the shape embedding for TMCAD.

\subsection{Two Graph-JEPA comparisons}
\label{sec:s-graphjepa}
The official Graph-JEPA implementation uses a two-layer GINE patch encoder, four-layer context and target transformers, width 512 and a three-layer predictor. To apply this to B-Reps, we combine the three entity types from our graphs (face, edge, vertex) into a homogeneous graph with typed edges and initialize 512-dimensional node embeddings using trainable versions of the geometric input encoders. We make the loop-order relation symmetric for partitioning. The configuration uses 32 patch slots. When a graph has fewer than 32 nodes, it assigns one node per nonempty slot. Larger graphs use METIS~\cite{karypis1998metis} after adjacency dropout at rate $0.3$. The configuration requests one context and four targets with patches expanded by one hop. Adam uses a learning rate of $5{\cdot}10^{-4}$ with a plateau scheduler, batch size 128 and a 30-epoch budget of 23,460 updates.

We noticed that the pretext loss overflowed in all seeds after 1,750, 5,550 and 7,650 updates. The target computes $(\cosh m,\sinh m)$ from an embedding mean $m$. We observed that patch activations grew by about a factor of 20 before overflow and that the target encoder's attention scaled with patch overlap. We enabled the row normalization option that was already present in the official implementation. This delayed overflow in all seeds. We use the last saved checkpoint common to the seeds with finite loss (epoch 13). For face segmentation, validation at seed 42 selected the concatenation of the face's mean patch-encoder state and its mean target-encoder patch output. For TMCAD it selected the official graph readout. \Cref{tab:s3-gj-official} gives the checkpoints' results.

\begin{table}[htbp]
  \centering
  \caption{\textbf{Official Graph-JEPA at the last common finite checkpoint.} Frozen-probe macro-F1 (\%) at epoch 13 of 30 (update 10,166) with row-normalized patch adjacency. Overflow gives the first update with infinite loss for that seed. The validation readout is selected at seed 42.}
  \label{tab:s3-gj-official}
  \footnotesize
  \setlength{\tabcolsep}{6pt}
  \begin{tabular}{@{}llcccccc@{}}
  \toprule
  & & \multicolumn{2}{c}{Fusion~360} & \multicolumn{2}{c}{MFInstSeg} & \multicolumn{2}{c}{TMCAD} \\
  \cmidrule(lr){3-4}\cmidrule(lr){5-6}\cmidrule(lr){7-8}
  Seed & Overflow at update & val & test & val & test & val & test \\
  \midrule
  42 & 19{,}450 & 53.54 & 52.92 & 61.72 & 61.29 & 74.41 & 73.28 \\
  1 & 12{,}900 & 55.30 & 54.97 & 63.98 & 64.17 & 69.92 & 70.43 \\
  2 & 10{,}650 & 65.70 & 64.46 & 75.27 & 74.60 & 77.00 & 76.96 \\
  \midrule
  Mean $\pm$ std & & $58.18{\pm}6.57$ & $57.45{\pm}6.16$ & $66.99{\pm}7.26$ & $66.69{\pm}7.00$ & $73.78{\pm}3.58$ & $73.55{\pm}3.27$ \\
  \bottomrule
  \end{tabular}
\end{table}

For the \emph{Graph-JEPA objective}, we use the same heterogeneous encoder as \method. We split the face-adjacency graph into 32 METIS patches after dropping each directed adjacency entry with probability $0.3$. When the graph has fewer than 32 faces, we start with one patch per face. The one-hop-expanded context patch is encoded as an induced subgraph. We mean-pool up to four expanded target patches from the full-solid EMA encoding and apply the hyperbolic target transformation to their pooled representations. The predictor uses max-pooled random-walk structural encodings ($k=20$), an MLP, and Smooth-L1. It uses a per-step EMA and batches of 16 solids in the main-paper configuration. \Cref{sec:s-graphjepa-schedules} gives the schedule comparison.

\subsection{Reconstruction with the shared encoder}
\label{sec:s-recon}

Same-corruption reconstruction uses the shared encoder. It uses the target sampler and mask construction of \method, including hidden end vertices and removed loop links. Its loss sums the mean squared coordinate error over valid face UV samples and the mean squared coordinate error over valid curve control points, using only masked faces or edges in the corresponding term. A face decoder projects to a $4{\times}4$ grid and then two transposed convolutions and an output convolution follow. Together, they add $2.07$\,M parameters during pretraining. This reconstruction baseline trains for the same number of updates as \method, with warm-up over the first $5\%$ of steps. Random initialization uses the same encoder without pretraining.

\FloatBarrier

\section{Transfer from frozen and fine-tuned encoders}
\label{sec:s-results}

The comparison with same-corruption reconstruction holds the encoder and masking rule fixed. The largest paired improvement is $9.19$ points on MFInstSeg, with a $95\%$ interval of $[8.72,9.69]$. On Fusion~360 the improvement is $3.66$ points, or $3.08$ when RevolveEnd, which accounts for 73 of the 77,070 test faces, is excluded from the macro-average (\cref{tab:s-paired}).

\begin{table}[htbp]
  \centering
  \caption{\textbf{Paired frozen performance gaps on the shared encoder.} Differences are \method minus the indicated objective, in points, with $95\%$ bootstrap intervals. F1$^\dagger$ excludes RevolveEnd. The other three BrepEDIT instruction styles give MRR gains of $1.2$--$1.4$ over same-corruption reconstruction and $3.8$--$5.7$ over the Graph-JEPA objective. All corresponding intervals exclude zero.}
  \label{tab:s-paired}
  \footnotesize
  \setlength{\tabcolsep}{6pt}
  \begin{tabular}{@{}llccc@{}}
    \toprule
    Task & Metric & $-$ same-corr.\ recon. & $-$ Graph-JEPA objective \\
    \midrule
    Fusion~360 & macro-F1 & $+3.66$ $[+1.94,+5.43]$ & $+2.61$ $[-0.15,+5.86]$ \\
    & F1$^\dagger$ (w/o RevolveEnd) & $+3.08$ $[+2.42,+3.78]$ & $+0.98$ $[-0.44,+2.69]$ \\
    MFInstSeg & macro-F1 & $+9.19$ $[+8.72,+9.69]$ & $+9.55$ $[+6.34,+15.48]$ \\
    TMCAD & macro-F1 & $+0.91$ $[-0.72,+2.60]$ & $+5.07$ $[+3.16,+7.27]$ \\
    \midrule
    BrepEDIT & MRR, geometric & $+1.3$ $[+1.1,+1.6]$ & $+5.2$ $[+4.5,+6.0]$ \\
    & R@1, geometric & $+1.8$ $[+1.4,+2.2]$ & $+7.1$ $[+6.1,+8.3]$ \\
    \bottomrule
  \end{tabular}
\end{table}

The Fusion~360 difference to the Graph-JEPA objective is uncertain across seeds: its interval is $[-0.15,+5.86]$ for macro-F1 and $[-0.44,+2.69]$ without RevolveEnd. On TMCAD, the $0.91$-point difference to same-corruption reconstruction has an interval that includes zero. The differences to the Graph-JEPA objective on MFInstSeg and TMCAD ($9.55$ and $5.07$ points) and all BrepEDIT differences have intervals that exclude zero.

\begin{table}[htbp]
  \centering
  \caption{\textbf{Frozen results by pretraining seed.} Scores for Fusion~360, MFInstSeg, and TMCAD are macro-F1 (\%). Geometric-instruction retrieval reports MRR and Recall@1 ($\times100$), each averaged over three head seeds. For Brep2Shape and UV-Net, the first column in each block gives the only pretraining run.}
  \label{tab:s-perseed}
  \footnotesize
  \setlength{\tabcolsep}{2.6pt}
  \begin{tabular}{@{}lccccccccccccccc@{}}
  \toprule
  & \multicolumn{3}{c}{Fusion~360} & \multicolumn{3}{c}{MFInstSeg} & \multicolumn{3}{c}{TMCAD} & \multicolumn{3}{c}{BrepEDIT MRR} & \multicolumn{3}{c}{BrepEDIT R@1} \\
  \cmidrule(lr){2-4}\cmidrule(lr){5-7}\cmidrule(lr){8-10}\cmidrule(lr){11-13}\cmidrule(l){14-16}
  Method & 42 & 1 & 2 & 42 & 1 & 2 & 42 & 1 & 2 & 42 & 1 & 2 & 42 & 1 & 2 \\
  \midrule
  Random initialization (reference) & 77.11 & 76.35 & 74.09 & 80.59 & 79.81 & 78.87 & 76.28 & 77.60 & 78.11 & 65.8 & 65.3 & 65.0 & 45.9 & 45.2 & 44.9 \\
  \midrule
  UV-Net & 35.05 & -- & -- & 21.46 & -- & -- & 69.48 & -- & -- & 61.6 & -- & -- & 39.9 & -- & -- \\
  Graph-JEPA & 52.92 & 54.97 & 64.46 & 61.29 & 64.17 & 74.60 & 73.28 & 70.43 & 76.96 & 62.5 & 60.9 & 61.9 & 41.3 & 39.1 & 40.4 \\
  Brep2Shape & 48.72 & -- & -- & 44.80 & -- & -- & 70.13 & -- & -- & 62.5 & -- & -- & 41.1 & -- & -- \\
  BRepMAE & 75.11 & 75.28 & 74.43 & 79.79 & 80.02 & 80.46 & 77.31 & 75.93 & 76.30 & 66.6 & 66.4 & 66.4 & 46.7 & 46.6 & 46.5 \\
  \midrule
  Same-corruption reconstruction & 78.69 & 77.54 & 76.66 & 80.75 & 80.83 & 79.71 & 78.75 & 79.26 & 79.49 & 67.8 & 67.7 & 67.7 & 48.6 & 48.6 & 48.6 \\
  Graph-JEPA objective & 81.02 & 75.09 & 79.93 & 83.07 & 74.46 & 82.68 & 76.35 & 73.08 & 75.58 & 64.5 & 63.2 & 64.1 & 44.2 & 42.2 & 43.5 \\
  \midrule
  \method & \hl 81.50 & \hl 80.84 & \hl 81.53 & \hl 89.48 & \hl 89.98 & \hl 89.38 & \hl 80.63 & \hl 79.72 & \hl 79.87 & \hl 69.1 & \hl 69.2 & \hl 68.9 & \hl 50.4 & \hl 50.6 & \hl 50.2 \\
  \bottomrule
  \end{tabular}
\end{table}

\Cref{tab:s-perseed} lists the individual pretraining seeds. \method has the highest score in every seed on all three probing benchmarks and on both BrepEDIT metrics. Its scores vary by at most $0.91$ points across seeds on the probing benchmarks, compared with $3.27$--$8.61$ points for the Graph-JEPA objective, which is the source of the wide intervals in that comparison. On Fusion~360, the margin over the Graph-JEPA objective ranges from $0.48$ to $5.75$ points across seeds.

\begin{table}[htbp]
  \centering
  \caption{\textbf{Additional frozen evaluation metrics on the shared encoder.} Values are averaged over three pretraining seeds. Accuracy, mean IoU, and Fusion~360 macro-F1 excluding RevolveEnd (F1$^\dagger$) are percentages. Effective rank is the exponential of the entropy of the normalized covariance spectrum over 20,000 frozen Fusion~360 test-face embeddings. Retrieval reports MRR ($\times100$) for the other three instruction styles.}
  \label{tab:s-secondary}
  \footnotesize
  \setlength{\tabcolsep}{4pt}
  \begin{tabular}{@{}lccccccccccc@{}}
  \toprule
  & \multicolumn{4}{c}{Fusion~360} & \multicolumn{2}{c}{MFInstSeg} & \multicolumn{2}{c}{TMCAD} & \multicolumn{3}{c}{BrepEDIT MRR} \\
  \cmidrule(lr){2-5}\cmidrule(lr){6-7}\cmidrule(lr){8-9}\cmidrule(l){10-12}
  Method & Acc. & mIoU & F1$^\dagger$ & Eff.\ rank & Acc. & mIoU & Acc. & mIoU & Detailed & Functional & Concise \\
  \midrule
  Random initialization (reference) & 86.28 & 64.47 & 82.30 & 22.5 & 87.20 & 69.43 & 77.84 & 64.30 & 68.3 & 64.7 & 63.8 \\
  Same-corruption reconstruction & 85.73 & 65.70 & 81.93 & 37.1 & 88.00 & 69.75 & 79.81 & 66.71 & 71.1 & 66.6 & 65.7 \\
  Graph-JEPA objective & \textbf{87.71} & 67.80 & 84.03 & 5.4 & 87.88 & 70.99 & 75.79 & 61.21 & 66.8 & 64.1 & 63.1 \\
  \method & \hl 87.46 & \hl \textbf{70.51} & \hl \textbf{85.01} & \hl 67.9 & \hl \textbf{94.34} & \hl \textbf{82.78} & \hl \textbf{80.65} & \hl \textbf{67.91} & \hl \textbf{72.5} & \hl \textbf{67.9} & \hl \textbf{66.9} \\
  \bottomrule
  \end{tabular}
\end{table}

\Cref{tab:s-secondary} reports accuracy, mean IoU, and retrieval for the other three instruction styles. \method has the highest value in every column except Fusion~360 accuracy, where the Graph-JEPA objective is $0.25$ points higher ($87.71$ versus $87.46$). Accuracy weights faces rather than classes. On mean IoU, which like macro-F1 weights the eight classes equally, \method leads the Graph-JEPA objective by $2.71$ points on Fusion~360. On MFInstSeg, \method improves mean IoU by $11.79$ points over the best other objective on the shared encoder ($82.78$ versus $70.99$). The effective rank of the frozen Fusion~360 face embeddings is $67.9$ for \method, compared with $37.1$ for same-corruption reconstruction and $5.4$ for the Graph-JEPA objective. On the detailed, functional, and concise instruction styles, \method leads same-corruption reconstruction by $1.2$--$1.4$ MRR points.

\subsection{Is the frozen representation useful with few labels?}
\label{sec:s-labeleff}
\begin{figure}[htbp]
  \centering
  \includegraphics[width=0.91\linewidth]{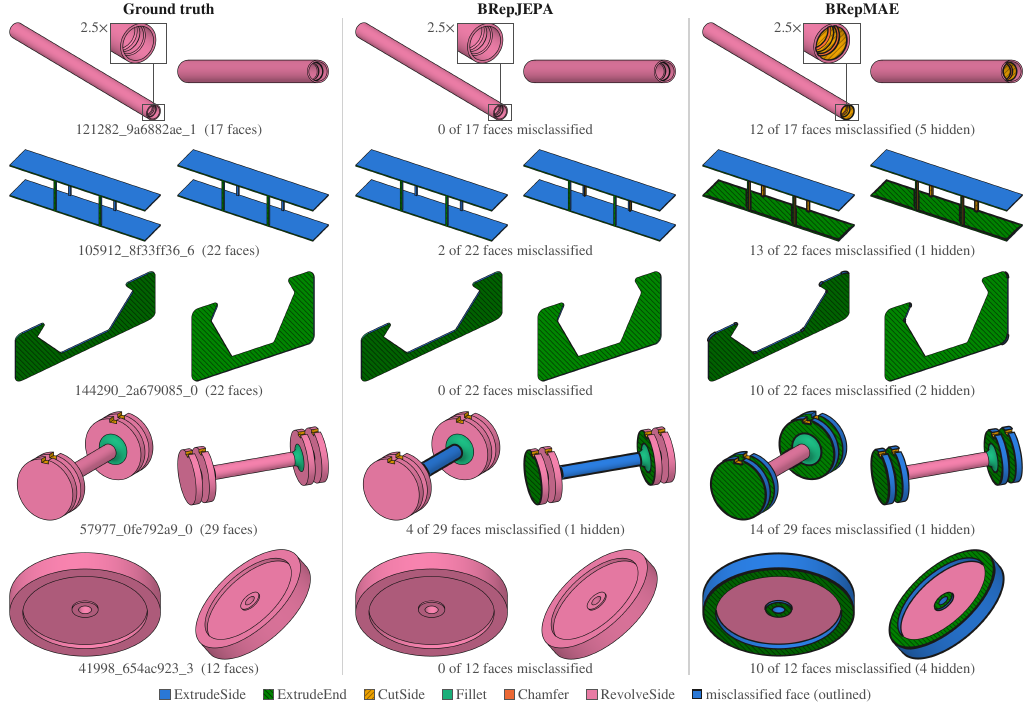}
  \caption{\textbf{Qualitative comparison with BRepMAE on five parts.} Ground truth and frozen linear-probe predictions (seed 42) of \method and BRepMAE on Fusion360 test parts, two views per part, misclassified faces outlined. \method correctly classifies $12$, $11$, $10$, $10$ and $10$ additional faces.}
  \label{fig:s3-qual-brepmae}
\end{figure}

The labeled training solids are fixed within each draw for methods on the shared encoder. \method improves over same-corruption reconstruction by $5.58$ points on Fusion~360 (212 solids), $18.76$ on MFInstSeg (80 solids) and $6.98$ on TMCAD (76 solids), at $1\%$ of the training labels. All three differences are positive in every benchmark and the gaps shrink as the training set grows. The Graph-JEPA objective on the same encoder has comparable Fusion~360 scores at small label fractions. \Cref{tab:s-labeleff-matched} presents paired differences for the shared encoder. \Cref{tab:s3-labeleff} gives the published-method scores plotted in \mainref{Fig.~5}. The closest method is BRepMAE on Fusion~360 and MFInstSeg, where the observed margins span $4.5$--$6.1$ and $9.5$--$11.3$ points across fractions, respectively. We present five qualitative examples from the Fusion 360 test set, comparing \method with BRepMAE in \Cref{fig:s3-qual-brepmae}.

\begin{table}[htbp]
  \centering
  \caption{\textbf{Frozen label efficiency on the shared encoder.} Each cell gives the macro-F1 gap in points (\method minus baseline). Three seed pairs share ten nested subsets per reduced fraction.}
  \label{tab:s-labeleff-matched}
  \footnotesize
  \setlength{\tabcolsep}{5pt}
\begin{tabular}{llccccc}
\toprule
 & \method $-$ & 1\% & 5\% & 10\% & 25\% & 100\% \\
\midrule
Fusion 360 & Graph-JEPA objective & $-0.06$\,{\scriptsize } & $-0.57$\,{\scriptsize } & $-0.04$\,{\scriptsize} & $+1.11$\,{\scriptsize } & $+1.82$\,{\scriptsize } \\
 & Same-corruption recon. & $+5.58$\,{\scriptsize } & $+4.53$\,{\scriptsize } & $+4.30$\,{\scriptsize } & $+3.53$\,{\scriptsize } & $+3.36$\,{\scriptsize } \\
 & Random initialization & $+4.83$\,{\scriptsize } & $+3.97$\,{\scriptsize } & $+4.11$\,{\scriptsize } & $+4.58$\,{\scriptsize } & $+5.19$\,{\scriptsize } \\
\midrule
MFInstSeg & Graph-JEPA objective & $+2.99$\,{\scriptsize } & $+6.32$\,{\scriptsize} & $+7.61$\,{\scriptsize} & $+8.58$\,{\scriptsize } & $+9.59$\,{\scriptsize } \\
 & Same-corruption recon. & $+18.76$\,{\scriptsize } & $+14.36$\,{\scriptsize } & $+12.21$\,{\scriptsize } & $+10.32$\,{\scriptsize } & $+9.15$\,{\scriptsize } \\
 & Random initialization & $+12.49$\,{\scriptsize } & $+11.90$\,{\scriptsize } & $+11.31$\,{\scriptsize } & $+10.31$\,{\scriptsize } & $+9.90$\,{\scriptsize } \\
\midrule
TMCAD & Graph-JEPA objective & $+12.92$\,{\scriptsize } & $+10.68$\,{\scriptsize } & $+8.70$\,{\scriptsize } & $+6.82$\,{\scriptsize } & $+5.09$\,{\scriptsize } \\
 & Same-corruption recon. & $+6.98$\,{\scriptsize } & $+4.23$\,{\scriptsize } & $+2.78$\,{\scriptsize } & $+1.23$\,{\scriptsize } & $+0.84$\,{\scriptsize } \\
 & Random initialization & $+10.63$\,{\scriptsize } & $+7.39$\,{\scriptsize } & $+5.56$\,{\scriptsize } & $+3.45$\,{\scriptsize } & $+2.74$\,{\scriptsize } \\
\bottomrule
\end{tabular}
\end{table}
\begin{table}[htbp]
  \centering
  \caption{\textbf{Frozen label efficiency with published methods.} Test macro-F1 (\%) for \mainref{Fig.~5}. For three-seed methods, entries summarize the per-seed means over three subset draws. For single-seed methods, standard deviations are across draws and the full-data cell is a single fit.}
  \label{tab:s3-labeleff}
  \footnotesize
  \setlength{\tabcolsep}{5pt}
  \begin{tabular}{@{}llcccccc@{}}
  \toprule
  Benchmark & Method & Seeds & $1\%$ & $5\%$ & $10\%$ & $25\%$ & $100\%$ \\
  \midrule
  Fusion~360 & \method & 3 & \hl {\boldmath$64.27{\pm}0.49$} & \hl {\boldmath$72.61{\pm}0.05$} & \hl {\boldmath$75.99{\pm}0.33$} & \hl {\boldmath$78.19{\pm}0.52$} & \hl {\boldmath$81.00{\pm}0.38$} \\
   & BRepMAE & 3 & $59.72{\pm}0.21$ & $67.94{\pm}0.50$ & $70.57{\pm}0.04$ & $73.06{\pm}0.08$ & $74.86{\pm}0.75$ \\
   & Graph-JEPA & 3 & $42.27{\pm}2.39$ & $51.28{\pm}3.79$ & $53.86{\pm}4.46$ & $56.09{\pm}5.87$ & $57.71{\pm}6.55$ \\
   & Brep2Shape & 1 & $40.36{\pm}2.64$ & $45.38{\pm}2.84$ & $47.12{\pm}1.68$ & $48.37{\pm}0.46$ & 48.62 \\
   & UV-Net & 1 & $27.06{\pm}0.80$ & $31.69{\pm}1.06$ & $33.70{\pm}0.62$ & $34.34{\pm}0.40$ & 35.05 \\
  \midrule
  MFInstSeg & \method & 3 & \hl {\boldmath$63.45{\pm}0.38$} & \hl {\boldmath$79.10{\pm}0.35$} & \hl {\boldmath$83.63{\pm}0.40$} & \hl {\boldmath$87.15{\pm}0.32$} & \hl {\boldmath$89.62{\pm}0.32$} \\
   & BRepMAE & 3 & $52.19{\pm}0.50$ & $67.75{\pm}0.52$ & $72.88{\pm}0.50$ & $77.01{\pm}0.36$ & $80.08{\pm}0.36$ \\
   & Graph-JEPA & 3 & $41.76{\pm}2.99$ & $53.86{\pm}4.41$ & $58.48{\pm}5.11$ & $62.90{\pm}6.00$ & $66.59{\pm}7.16$ \\
   & Brep2Shape & 1 & $27.39{\pm}0.29$ & $37.54{\pm}0.19$ & $40.74{\pm}0.21$ & $43.11{\pm}0.19$ & 44.91 \\
   & UV-Net & 1 & $12.43{\pm}0.20$ & $17.45{\pm}0.11$ & $19.36{\pm}0.21$ & $20.77{\pm}0.10$ & 21.46 \\
  \midrule
  TMCAD & \method & 3 & \hl {\boldmath$46.08{\pm}1.00$} & \hl {\boldmath$66.29{\pm}0.53$} & \hl {\boldmath$69.97{\pm}0.71$} & \hl {\boldmath$74.90{\pm}0.30$} & \hl {\boldmath$80.06{\pm}0.46$} \\
   & BRepMAE & 3 & $35.02{\pm}0.07$ & $56.77{\pm}0.78$ & $62.33{\pm}0.84$ & $70.43{\pm}0.14$ & $76.51{\pm}0.70$ \\
   & Graph-JEPA & 3 & $38.88{\pm}2.15$ & $57.97{\pm}1.80$ & $62.80{\pm}1.64$ & $68.41{\pm}2.19$ & $73.56{\pm}3.41$ \\
   & Brep2Shape & 1 & $35.07{\pm}4.71$ & $54.73{\pm}0.89$ & $59.17{\pm}1.05$ & $64.76{\pm}1.15$ & 70.13 \\
   & UV-Net & 1 & $31.32{\pm}4.98$ & $53.69{\pm}2.58$ & $60.15{\pm}1.08$ & $65.83{\pm}0.07$ & 69.48 \\
  \bottomrule
  \end{tabular}
\end{table}

\subsection{How much of the gap survives fine-tuning?}
\label{sec:s-finetune}
With full fine-tuning and all available labels, the compared objectives on the shared encoder are within $0.48$ points of one another on Fusion~360 and $0.16$ points on MFInstSeg ((\cref{tab:s-finetune}). On scarce labels, \method preserves some of the frozen advantage. At $1\%$ on Fusion~360, the difference to same-corruption reconstruction is $+1.52$ points. At $5\%$ and $10\%$ on MFInstSeg the intervals favor \method over all shared-encoder pretraining baselines (\cref{tab:s-fewshot}).

BRepMAE fine-tunes better than \method on MFInstSeg. The seed-42 BRepMAE architecture trained from random initialization reaches $80.50$, $94.14$, and $95.81$ at $1$, $5$, and $10\%$ of the labels. The latter two scores already exceed the corresponding \method means ($93.17$ and $95.15$). \Cref{tab:s3-ft-brepmae} reports BRepMAE fine-tuning results with both pretrained and randomly initialized weights. On Fusion~360, the main-paper comparison favors \method over BRepMAE at every fraction. 

\begin{table}[htbp]
  \centering
  \caption{\textbf{End-to-end fine-tuning at the full label budget.} Test macro-F1 (\%) by pretraining seed, with validation-selected learning rates and zero-based epoch indices. Paired gaps have $95\%$ hierarchical intervals. F1$^\dagger$ excludes RevolveEnd. MFInstSeg uses the fixed 8,000-solid training subset.}
  \label{tab:s-finetune}
  \footnotesize
  \setlength{\tabcolsep}{3.5pt}
  \begin{tabular}{@{}llcccccccc@{}}
    \toprule
    & Method & lr & Seed 42 & Seed 1 & Seed 2 & Mean $\pm$ std & F1$^\dagger$ & Best epoch & \method $-$ X \\
    \midrule
    \multirow{4}{*}{\rotatebox{90}{MFInstSeg}}
    & Random init.\ (ref.) & $5{\cdot}10^{-4}$ & 98.27 & 98.11 & 98.29 & $98.22{\pm}0.10$ & -- & 33 / 26 / 45 & -- \\
    & Same-corr.\ recon. & $5{\cdot}10^{-4}$ & 98.35 & 98.28 & 97.95 & $98.19{\pm}0.21$ & -- & 40 / 35 / 20 & $+0.16$ {\scriptsize$[-0.32,+0.55]$} \\
    & Graph-JEPA objective & $5{\cdot}10^{-4}$ & 98.27 & 98.26 & 98.24 & $98.26{\pm}0.01$ & -- & 38 / 46 / 40 & $+0.10$ {\scriptsize$[-0.23,+0.33]$} \\
    & \method & \hl $5{\cdot}10^{-4}$ & \hl 98.01 & \hl 98.53 & \hl 98.53 & \hl $98.36{\pm}0.30$ & \hl -- & \hl 12 / 46 / 44 & \hl -- \\
    \midrule
    \multirow{4}{*}{\rotatebox{90}{Fusion~360}}
    & Random init.\ (ref.) & $5{\cdot}10^{-4}$ & 90.27 & 91.11 & 91.42 & $90.93{\pm}0.59$ & $93.89{\pm}0.27$ & 30 / 32 / 31 & -- \\
    & Same-corr.\ recon. & $5{\cdot}10^{-4}$ & 91.16 & 91.48 & 90.70 & $91.11{\pm}0.39$ & $94.56{\pm}0.11$ & 34 / 31 / 23 & $+0.36$ {\scriptsize$[-0.84,+1.55]$} \\
    & Graph-JEPA objective & $5{\cdot}10^{-4}$ & 91.07 & 91.01 & 90.89 & $90.99{\pm}0.09$ & $93.70{\pm}0.19$ & 25 / 33 / 29 & $+0.48$ {\scriptsize$[-0.64,+1.56]$} \\
    & \method & \hl $10^{-4}$ & \hl 91.20 & \hl 91.61 & \hl 91.60 & \hl $91.47{\pm}0.23$ & \hl $94.17{\pm}0.06$ & \hl 31 / 36 / 22 & \hl -- \\
    \bottomrule
  \end{tabular}
\end{table}
\begin{table}[htbp]
  \centering
  \caption{\textbf{End-to-end fine-tuning with few labels.} Macro-F1 gaps in points (\method minus baseline) with $95\%$ hierarchical intervals. Each gap averages three seed pairs and three subset draws. F1$^\dagger$ excludes RevolveEnd.}
  \label{tab:s-fewshot}
  \scriptsize
  \setlength{\tabcolsep}{2pt}
  \begin{tabular}{@{}llccc@{}}
    \toprule
    Benchmark & \method $-$ X & $1\%$ & $5\%$ & $10\%$ \\
    \midrule
    MFInstSeg & Same-corr.\ recon. & $+2.01$ {\scriptsize$[+0.97,+3.20]$} & $+1.20$ {\scriptsize$[+0.90,+1.62]$} & $+0.79$ {\scriptsize$[+0.47,+1.14]$} \\
     & Graph-JEPA objective & $+0.76$ {\scriptsize$[-0.26,+1.92]$} & $+1.22$ {\scriptsize$[+0.95,+1.50]$} & $+1.03$ {\scriptsize$[+0.89,+1.19]$} \\
    \midrule
    Fusion~360 & Same-corr.\ recon. & $+1.52$ {\scriptsize$[+0.53,+2.46]$} & $+0.73$ {\scriptsize$[-0.37,+1.89]$} & $+0.34$ {\scriptsize$[-0.66,+1.36]$} \\
     & Graph-JEPA objective & $+1.75$ {\scriptsize$[+0.46,+3.04]$} & $+2.38$ {\scriptsize$[+1.07,+3.64]$} & $+2.02$ {\scriptsize$[+0.86,+3.28]$} \\
    \midrule
    Fusion~360, F1$^\dagger$ & Same-corr.\ recon. & $+1.82$ {\scriptsize$[+1.15,+2.48]$} & $+0.37$ {\scriptsize$[-0.21,+0.99]$} & $+0.12$ {\scriptsize$[-0.37,+0.62]$} \\
     & Graph-JEPA objective & $+2.07$ {\scriptsize$[+1.01,+3.25]$} & $+2.26$ {\scriptsize$[+1.50,+3.16]$} & $+1.62$ {\scriptsize$[+1.12,+2.17]$} \\
    \bottomrule
  \end{tabular}
\end{table}
\begin{table}[htbp]
  \centering
  \caption{\textbf{Effect of initialization on BRepMAE fine-tuning.} Test macro-F1 (\%). For three-seed rows, standard deviations are across seed means. For one-seed rows, they are across subset draws.}
  \label{tab:s3-ft-brepmae}
  \scriptsize
  \setlength{\tabcolsep}{1.1pt}
  \begin{tabular}{@{}lcccccccc@{}}
  \toprule
  & \multicolumn{4}{c}{Fusion~360} & \multicolumn{4}{c}{MFInstSeg} \\
  \cmidrule(lr){2-5}\cmidrule(l){6-9}
  Initialization & $1\%$ & $5\%$ & $10\%$ & $100\%$ & $1\%$ & $5\%$ & $10\%$ & $100\%$ \\
  \midrule
  BRepMAE$^{\dagger}$ (3 seeds) & $67.27{\pm}0.45$ & $77.49{\pm}0.20$ & $80.37{\pm}0.44$ & $87.79{\pm}0.27$ & $82.05{\pm}0.67$ & $94.71{\pm}0.16$ & $96.16{\pm}0.10$ & $98.73{\pm}0.14$ \\
  BRepMAE, random init.\ (1 seed) & $64.85{\pm}2.27$ & -- & -- & -- & $80.50{\pm}1.02$ & $94.14{\pm}0.21$ & $95.81{\pm}0.30$ & {\boldmath$98.74$} \\
  \midrule
  \method (3 seeds) & \hl {\boldmath$71.73{\pm}0.33$} & \hl {\boldmath$82.93{\pm}0.12$} & \hl {\boldmath$85.67{\pm}0.12$} & \hl {\boldmath$91.47{\pm}0.23$} & \hl {\boldmath$82.40{\pm}0.38$} & \hl $93.17{\pm}0.13$ & \hl $95.15{\pm}0.09$ & \hl $98.36{\pm}0.30$ \\
  \bottomrule
  \end{tabular}
\end{table}

\FloatBarrier

\section{What contributes to transfer?}
\label{sec:s-ablations}

\subsection{Teacher depth}
The final-layer control uses the normalized layer 4 as its target and keeps the same normalization. When we average layers 0--4 instead, we observe an improvement in macro-F1 over three seeds of $3.63$ points on Fusion~360 and $5.50$ on MFInstSeg (\cref{tab:s3-three}). In the seed-42 depth sweep, layers $\{0,4\}$ give a similar mean validation score to the full average, while omitting the earliest layers, as in $\{2,3,4\}$, lowers it by $1.60$ points (\cref{tab:s-depth}). Based on these results, we include early layers in the target. The geometric probes in \cref{sec:s-geometric} measure which information is accessible in the frozen representation.

\begin{table}[htbp]
  \centering
  \caption{\textbf{Ablations repeated across three seeds.} Frozen-probe macro-F1 (\%). Validation averages over seeds and also gives the mean and sample standard deviation of the per-seed means over tasks. $\Delta$ is the validation mean minus the reference mean, in points. Test gives mean $\pm$ standard deviation across seeds.}
  \label{tab:s3-three}
  \footnotesize
  \setlength{\tabcolsep}{3.5pt}
  \begin{tabular}{@{}lcccccccc@{}}
  \toprule
  & \multicolumn{5}{c}{Validation} & \multicolumn{3}{c}{Test} \\
  \cmidrule(lr){2-6}\cmidrule(lr){7-9}
  Variant & Fusion~360 & MFInstSeg & TMCAD & Mean & $\Delta$ & Fusion~360 & MFInstSeg & TMCAD \\
  \midrule
  \method & \hl \textbf{80.82} & \hl 89.60 & \hl 79.87 & \hl {\boldmath$83.43{\pm}0.16$} & \hl -- & \hl {\boldmath$81.29{\pm}0.39$} & \hl $89.61{\pm}0.32$ & \hl {\boldmath$80.08{\pm}0.49$} \\
  \midrule
  \multicolumn{9}{@{}l}{\emph{Teacher targets}} \\
  Final teacher layer only & 76.97 & 84.15 & 78.66 & $79.93{\pm}0.12$ & $-3.50$ & $77.66{\pm}0.42$ & $84.11{\pm}0.30$ & $79.00{\pm}0.48$ \\
  \midrule
  \multicolumn{9}{@{}l}{\emph{Target design}} \\
  Random local targets of matched size & 78.59 & 88.75 & \textbf{80.00} & $82.45{\pm}0.41$ & $-0.98$ & $79.82{\pm}0.41$ & $88.95{\pm}0.41$ & $79.45{\pm}0.82$ \\
  Half as many targets & 78.27 & 89.76 & 79.63 & $82.55{\pm}0.31$ & $-0.87$ & $79.72{\pm}0.30$ & $89.55{\pm}0.06$ & $79.17{\pm}0.11$ \\
  \midrule
  \multicolumn{9}{@{}l}{\emph{Predictor query}} \\
  No entity tokens in the query & 79.84 & 89.90 & 79.43 & $83.06{\pm}0.36$ & $-0.37$ & $80.87{\pm}0.60$ & {\boldmath$89.83{\pm}0.59$} & $79.52{\pm}0.96$ \\
  \midrule
  \multicolumn{9}{@{}l}{\emph{Regularizer}} \\
  VICReg instead of SIGReg & 78.16 & 89.69 & 79.68 & $82.51{\pm}0.27$ & $-0.92$ & $79.37{\pm}0.69$ & $89.76{\pm}0.29$ & $80.00{\pm}0.66$ \\
  No regularizer & 78.76 & \textbf{89.98} & 79.29 & $82.68{\pm}0.43$ & $-0.75$ & $79.72{\pm}0.58$ & $89.81{\pm}0.57$ & $79.79{\pm}0.55$ \\
  \bottomrule
  \end{tabular}
\end{table}

\begin{table}[htbp]
  \centering
  \caption{\textbf{Teacher-depth sensitivity at seed 42.} Mean validation macro-F1 across the three probes and test macro-F1 for each task, in percent. Every layer is normalized before target construction.}
  \label{tab:s-depth}
  \small
  \begin{tabular}{@{}lcccc@{}}
    \toprule
    Teacher target & Mean val. & Fusion~360 & MFInstSeg & TMCAD \\
    \midrule
    Final layer $\{4\}$ & 79.87 & 77.64 & 84.03 & 78.81 \\
    Layers $\{2,3,4\}$ & 81.66 & 80.11 & 87.68 & 79.75 \\
    Layers $\{1,2,3,4\}$ & 82.67 & 80.62 & 89.06 & 79.62 \\
    Layers $\{0,4\}$ & 83.36 & 80.44 & 89.63 & 80.08 \\
    Concatenated depths & 83.41 & 81.20 & 89.58 & 80.21 \\
    Mean of layers $\{0,1,2,3,4\}$ & 83.26 & 81.50 & 89.48 & 80.63 \\
    \bottomrule
  \end{tabular}
\end{table}

\subsection{Topology-defined targets and mask size}
\label{sec:s-targets}
We test the choice of loop and region-boundary edge sets at a matched masking budget against random local control sets. We replace each set by the same number of distinct edges drawn near the original anchor, using faces within one hop for all loops and within the region radius plus one hop for regions. We reject draws that recover the original set. The sampler matches the overlap with edges already masked by other targets so the masked-edge count is preserved. Coedges are kept and face draws match whether the sampled face was already masked. Both conditions hide every endpoint of a masked edge. On 1,000 MFInstSeg graphs the random sets hide more vertices ($30.0$ vs.\ $23.4$ per solid), and their loop edges lie an average $0.76$ face hops from the anchor rather than at the anchor.

The topology-defined targets yield higher test macro-F1 in all three seed pairs on each benchmark. We observe improvements of $1.47$ points on Fusion~360, $0.66$ on MFInstSeg ($[0.41,0.91]$) and $0.63$ on TMCAD ($[-0.70,+1.96]$). These gains measure the effect of using the topology-defined construction as a whole relative to this control. 

At $1\%$ of the labeled solids, our topology-defined targets outperform the random local controls by $1.92$ points on Fusion~360 and $1.98$ on MFInstSeg. Each of the 30 differences from the three paired seeds and ten subset draws is positive on both benchmarks. In \Cref{tab:s-targets-labeleff}, we report the gaps across label fractions. Specifically, as it is shown in \Cref{tab:s-targets-perclass}, on Fusion~360 the gains are most consistent for the cut and extrude classes and for RevolveEnd. CutSide and CutEnd gain $3.69$ and $3.44$ points and RevolveEnd gains $2.70$, while Fillet, RevolveSide, and Chamfer change by less than $0.4$ points. On MFInstSeg, 22 of 25 classes are higher in every seed, while the rectangular through slot is lower in every seed ($-0.91$ points on average).

\begin{table}[htbp]
  \centering
  \caption{\textbf{Topology-defined and random local targets at reduced label budgets.} Macro-F1 gaps in points (\method minus random local), with $95\%$ hierarchical intervals.}
  \label{tab:s-targets-labeleff}
  \footnotesize
  \setlength{\tabcolsep}{5pt}
\begin{tabular}{lccccc}
\toprule
\method $-$ random local & 1\% & 5\% & 10\% & 25\% & 100\% \\
\midrule
Fusion~360 & $+1.92$\,{\scriptsize } & $+1.52$\,{\scriptsize} & $+1.42$\,{\scriptsize } & $+1.11$\,{\scriptsize } & $+1.17$\,{\scriptsize } \\
 & {\scriptsize$[+1.14,+2.75]$} & {\scriptsize$[+0.81,+2.14]$} & {\scriptsize$[+0.37,+2.39]$} & {\scriptsize$[+0.21,+1.97]$} & {\scriptsize$[-0.23,+2.69]$} \\
\midrule
MFInstSeg & $+1.98$\,{\scriptsize} & $+1.04$\,{\scriptsize } & $+0.94$\,{\scriptsize } & $+0.70$\,{\scriptsize } & $+0.66$\,{\scriptsize } \\
 & {\scriptsize$[+1.57,+2.36]$} & {\scriptsize$[+0.82,+1.28]$} & {\scriptsize$[+0.66,+1.31]$} & {\scriptsize$[+0.45,+0.98]$} & {\scriptsize$[+0.39,+0.91]$} \\
\midrule
TMCAD & $+0.37$\,{\scriptsize } & $+0.73$\,{\scriptsize } & $+0.70$\,{\scriptsize } & $+0.17$\,{\scriptsize } & $+0.63$\,{\scriptsize } \\
 & {\scriptsize$[-0.83,+1.60]$} & {\scriptsize$[-0.23,+1.66]$} & {\scriptsize$[-0.72,+2.05]$} & {\scriptsize$[-0.74,+1.09]$} & {\scriptsize$[-0.66,+1.97]$} \\
\bottomrule
\end{tabular}
\end{table}
\begin{table}[htbp]
  \begin{minipage}[t]{0.36\linewidth}
  \centering
  \caption{\textbf{Topology-defined and random local targets by Fusion~360 class.} Frozen-probe F1 (\%) averaged over three pretraining seeds. $\Delta$ is the mean paired gap in points.}
  \label{tab:s-targets-perclass}
  \footnotesize
  \setlength{\tabcolsep}{2.8pt}
  \begin{tabular}{@{}lcccc@{}}
    \toprule
    Class & Random local & \method & $\Delta$ \\
    \midrule
    ExtrudeSide & 89.7 & 90.6 & $+0.92$ \\
    ExtrudeEnd & 88.5 & 90.0 & $+1.46$ \\
    CutSide & 63.5 & 67.2 & $+3.69$ \\
    Fillet & 97.3 & 97.3 & $-0.05$ \\
    RevolveSide & 80.3 & 80.3 & $-0.07$ \\
    Chamfer & 92.7 & 92.4 & $-0.35$ \\
    CutEnd & 74.0 & 77.5 & $+3.44$ \\
    RevolveEnd & 52.5 & 55.2 & $+2.70$ \\
    \bottomrule
  \end{tabular}
  \end{minipage}\hfill
  \begin{minipage}[t]{0.60\linewidth}
  \centering
  \caption{\textbf{Topology-defined and random local targets at smaller pretraining dataset sizes.} Frozen-probe macro-F1 (\%) at seed 42. The pretraining sets are nested random subsets of the 100k solids, and every run performs about 362k updates. $\Delta$ is \method minus random local, in points.}
  \label{tab:s-targets-corpus}
  \footnotesize
  \setlength{\tabcolsep}{8pt}
  \begin{tabular}{@{}llcccc@{}}
  \toprule
  & & \multicolumn{2}{c}{Fusion~360} & \multicolumn{2}{c}{MFInstSeg} \\
  \cmidrule(lr){3-4}\cmidrule(lr){5-6}
  Solids & Targets & val & test & val & test \\
  \midrule
  1k & \method & \hl 78.29 & \hl 79.82 & \hl 86.45 & \hl 86.79 \\
   & random local & 77.14 & 77.92 & 86.00 & 86.02 \\
   & $\Delta$ & $+1.15$ & $+1.90$ & $+0.45$ & $+0.77$ \\
  \midrule
  10k & \method & \hl 80.90 & \hl 82.30 & \hl 89.39 & \hl 89.32 \\
   & random local & 78.98 & 79.96 & 88.10 & 88.23 \\
   & $\Delta$ & $+1.93$ & $+2.35$ & $+1.29$ & $+1.09$ \\
  \midrule
  30k & \method & \hl 80.73 & \hl 81.93 & \hl 89.07 & \hl 89.02 \\
   & random local & 79.13 & 79.19 & 88.67 & 88.71 \\
   & $\Delta$ & $+1.60$ & $+2.74$ & $+0.40$ & $+0.32$ \\
  \midrule
  100k & \method & \hl 80.93 & \hl 81.50 & \hl 89.18 & \hl 89.48 \\
   & random local & 78.13 & 80.30 & 88.45 & 89.03 \\
   & $\Delta$ & $+2.80$ & $+1.20$ & $+0.73$ & $+0.45$ \\
  \bottomrule
  \end{tabular}
  \end{minipage}
\end{table}

We repeat the comparison at seed 42 on nested random subsets of the pretraining dataset with 1k, 10k and 30k solids. At every dataset size, the topology-defined targets yield higher test macro-F1 on Fusion~360 and MFInstSeg. Results are presented in \Cref{tab:s-targets-corpus}. Each cell is a single run. 


Additionally, we perform two further controls. We change the number of targets per solid, and with it the share of hidden entities, which is $21.7\%$ of faces and $22.0\%$ of edges in \method. Reducing target counts from $2/4/1/3$ to $1/2/1/2$ masks $13.5\%$ of faces and $17.2\%$ of edges and lowers test macro-F1 by $1.57$ points on Fusion~360 and $0.90$ on TMCAD. Doubling the counts to $4/8/2/6$ masks $35.8\%$ of faces and $34.7\%$ of edges and changes the seed-42 mean validation score by $-0.26$ points.

\subsection{How much does the query identify?}
Removing the hidden entity tokens leaves the anchor, family, and orientation as the input to the predictor together with \textsc{cls}. This leads to a $-0.42$ change in frozen test macro-F1 relative to the full query on Fusion~360, $+0.22$ on MFInstSeg, and $-0.55$ on TMCAD. Entity tokens distinguish multiple targets having the same anchor, family, and orientation. This experiment leaves their transfer benefit unresolved. The comparison is paired by seed, with different data and mask streams due to the change in model initialization.

\subsection{Regularization and dimensionality}
\label{sec:s-regularization}
\begin{figure}[htbp]
  \centering
  \includegraphics[width=0.76\linewidth]{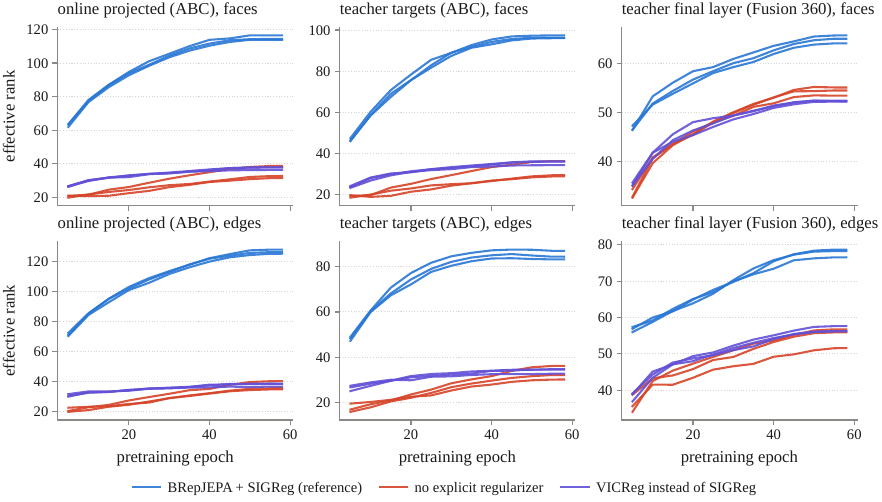}
  \caption{\textbf{Effective rank throughout pretraining.} Each line follows one seed. The first two columns show online projections and teacher targets on ABC. The last column shows the frozen teacher readout on Fusion~360. The top row gives face embeddings and the bottom row gives edge embeddings.}
  \label{fig:s-rank}
\end{figure}

We compare SIGReg with VICReg to assess the effect of regularization. VICReg uses a per-family variance term with target standard deviation $1$ and weight $1$, and a covariance term with weight $0.04$, on the predictions. On test, SIGReg outperforms VICReg by $1.92$ points on Fusion~360, while differences on the other benchmarks are smaller. Removing SIGReg lowers Fusion~360 macro-F1 by $1.57$ points, with a $95\%$ interval of $[0.10,3.09]$ for the drop. SIGReg also substantially increases the effective rank of the learned representations compared with VICReg and no regularization. \Cref{tab:s-regularizer,tab:s-noreg,fig:s-rank} report the downstream scores and effective ranks.

\begin{table}[htbp]
  \begin{minipage}[t]{0.44\linewidth}
  \centering
  \caption{\textbf{SIGReg and prediction-level VICReg.} Mean $\pm$ standard deviation over three pretraining seeds. Probe and fine-tuning (FT) scores use macro-F1 (\%). BrepEDIT MRR is multiplied by $100$. The last column gives paired differences with $95\%$ hierarchical intervals.}
  \label{tab:s-regularizer}
  \scriptsize
  \setlength{\tabcolsep}{2pt}
  \begin{tabular}{@{}lccc@{}}
    \toprule
    Task & SIGReg & VICReg & Difference \\
    \midrule
    Fusion~360 & $81.29{\pm}0.39$ & $79.37{\pm}0.69$ & $+1.92$ {\scriptsize$[+0.42,+3.43]$} \\
    MFInstSeg & $89.61{\pm}0.32$ & $89.76{\pm}0.29$ & $-0.14$ {\scriptsize$[-0.30,+0.02]$} \\
    TMCAD & $80.08{\pm}0.49$ & $80.00{\pm}0.66$ & $+0.07$ {\scriptsize$[-1.52,+1.64]$} \\
    BrepEDIT MRR & $69.1{\pm}0.1$ & $69.3{\pm}0.0$ & $-0.2$ {\scriptsize$[-0.4,-0.0]$} \\
    FT Fusion~360 & $91.47{\pm}0.23$ & $91.99{\pm}0.26$ & $-0.52$ {\scriptsize$[-1.54,+0.36]$} \\
    FT MFInstSeg & $98.36{\pm}0.30$ & $98.34{\pm}0.26$ & $+0.01$ {\scriptsize$[-0.07,+0.08]$} \\
    \bottomrule
  \end{tabular}
  \end{minipage}\hfill
  \begin{minipage}[t]{0.53\linewidth}
  \centering
  \caption{\textbf{Transfer and effective rank without SIGReg.} Frozen test macro-F1 (\%) is averaged over three seeds. Rank is computed from projected online faces on ABC and from the final teacher layer on Fusion~360 test faces. The gap is the unregularized run minus SIGReg with a $95\%$ hierarchical interval.}
  \label{tab:s-noreg}
  \scriptsize
  \setlength{\tabcolsep}{2.5pt}
  \begin{tabular}{@{}lcccc@{}}
    \toprule
    & SIGReg & None & VICReg & None $-$ SIGReg \\
    \midrule
    Fusion~360 & $81.29{\pm}0.39$ & $79.72{\pm}0.58$ & $79.37{\pm}0.69$ & $-1.57$ {\scriptsize$[-3.09,-0.10]$} \\
    MFInstSeg & $89.61{\pm}0.32$ & $89.81{\pm}0.57$ & $89.76{\pm}0.29$ & $+0.20$ {\scriptsize$[-0.43,+0.95]$} \\
    TMCAD & $80.08{\pm}0.49$ & $79.79{\pm}0.55$ & $80.00{\pm}0.66$ & $-0.28$ {\scriptsize$[-1.58,+0.98]$} \\
    \midrule
    Eff.\ rank, online & $115$ & $34$ & $37$ & \\
    Eff.\ rank, teacher & $65$ & $54$ & $52$ & \\
    \bottomrule
  \end{tabular}
  \end{minipage}
\end{table}
\FloatBarrier

\section{What the embeddings encode}
\label{sec:s-further}

\subsection{Local geometry and relational properties}
\label{sec:s-geometric}
To test local geometry, we fit linear probes on the frozen final-layer face and edge embeddings. The probes for edge convexity and cylindrical bosses versus holes use properties inferred from the geometry. The other probes use linearly accessible input attributes, such as face area, centroid, mean normal, and curve type. Among the encoders in \cref{tab:s-geom}, \method achieves the highest scores on both inferred properties on both benchmarks. Random local targets reduce the two scores from $0.881$ to $0.857$ and from $0.936$ to $0.914$, respectively. The changes on MFInstSeg are smaller ($0.965$ to $0.962$ and $0.982$ to $0.977$).

Some explicit attributes are retained more accurately by raw descriptors and reconstruction embeddings. In particular, reconstruction can make area and centroid more linearly accessible than \method. The final-layer target control at seed 42 has a lower mean-normal $R^2$ on MFInstSeg than the multi-depth variant ($0.556$ vs.\ a three-seed mean of $0.941$). The two inferred properties have Spearman correlations of $0.84$ and $0.91$ with MFInstSeg macro-F1, across 18 encoders. The boss/hole target is derived from the UV grid and correlated with the segmentation labels, so these correlations describe the probes and their tasks.

\begin{table}[htbp]
  \centering
  \caption{\textbf{Geometric attributes recoverable from frozen embeddings.} Convexity, boss/hole, and surface and curve type use balanced accuracy. Log-area, centroid, mean normal, and edge log-length use $R^2$. Standard deviations, where shown, are across three seeds. The final-layer target ablation uses seed 42.}
  \label{tab:s-geom}
  \scriptsize
  \setlength{\tabcolsep}{1.6pt}
  \textbf{MFInstSeg}\\[2pt]
\resizebox{0.97\textwidth}{!}{%
\begin{tabular}{lccccccc}
\toprule
 & E1 conv. & F1 boss & F3 area & F3 cent. & F3 norm. & E2 type & E2 len. \\
 & inf. & inf. & ret. & ret. & ret. & ret. & ret. \\
\midrule
random init & 0.930{\tiny$\pm$}0.001 & 0.960{\tiny$\pm$}0.002 & 0.843{\tiny$\pm$}0.011 & 0.956{\tiny$\pm$}0.006 & 0.985{\tiny$\pm$}0.003 & 0.998{\tiny$\pm$}0.000 & 0.978{\tiny$\pm$}0.009 \\
same-corruption recon. & 0.960 & 0.958 & 0.931 & 0.949 & 0.881 & 0.997 & 0.984 \\
\method, final layer only & 0.947 & 0.966 & 0.866 & 0.799 & 0.556 & 0.998 & 0.952 \\
\method, random local targets & 0.962{\tiny$\pm$}0.002 & 0.977{\tiny$\pm$}0.004 & 0.907{\tiny$\pm$}0.002 & 0.871{\tiny$\pm$}0.010 & 0.932{\tiny$\pm$}0.004 & 0.999{\tiny$\pm$}0.000 & 0.971{\tiny$\pm$}0.002 \\
\method & \hl 0.965{\tiny$\pm$}0.000 & \hl 0.982{\tiny$\pm$}0.000 & \hl 0.903{\tiny$\pm$}0.003 & \hl 0.894{\tiny$\pm$}0.000 & \hl 0.941{\tiny$\pm$}0.005 & \hl 0.999{\tiny$\pm$}0.000 & \hl 0.966{\tiny$\pm$}0.005 \\
\midrule
raw input features & 0.698 & 0.803 & 1.000 & 0.991 & 1.000 & 0.997 & 1.000 \\
\bottomrule
\end{tabular}}
  \\[6pt]\textbf{Fusion~360}\\[2pt]
\resizebox{0.97\textwidth}{!}{%
\begin{tabular}{lcccccccc}
\toprule
 & E1 conv. & F1 boss & F2 type & F3 area & F3 cent. & F3 norm. & E2 type & E2 len. \\
 & inf. & inf. & ret. & ret. & ret. & ret. & ret. & ret. \\
\midrule
random init & 0.832{\tiny$\pm$}0.008 & 0.868{\tiny$\pm$}0.005 & 0.996{\tiny$\pm$}0.001 & 0.744{\tiny$\pm$}0.006 & 0.942{\tiny$\pm$}0.006 & 0.937{\tiny$\pm$}0.009 & 0.999{\tiny$\pm$}0.000 & 0.962{\tiny$\pm$}0.016 \\
same-corruption recon. & 0.866 & 0.891 & 0.967 & 0.859 & 0.911 & 0.858 & 0.998 & 0.968 \\
\method, final layer only & 0.823 & 0.844 & 0.977 & 0.767 & 0.725 & 0.566 & 0.999 & 0.921 \\
\method, random local targets & 0.857{\tiny$\pm$}0.004 & 0.914{\tiny$\pm$}0.020 & 0.995{\tiny$\pm$}0.000 & 0.846{\tiny$\pm$}0.007 & 0.828{\tiny$\pm$}0.015 & 0.901{\tiny$\pm$}0.006 & 1.000{\tiny$\pm$}0.000 & 0.956{\tiny$\pm$}0.002 \\
\method & \hl 0.881{\tiny$\pm$}0.003 & \hl 0.936{\tiny$\pm$}0.001 & \hl 0.992{\tiny$\pm$}0.001 & \hl 0.828{\tiny$\pm$}0.012 & \hl 0.851{\tiny$\pm$}0.001 & \hl 0.912{\tiny$\pm$}0.005 & \hl 0.999{\tiny$\pm$}0.000 & \hl 0.938{\tiny$\pm$}0.005 \\
\midrule
raw input features & 0.655 & 0.711 & 1.000 & 1.000 & 0.996 & 0.999 & 1.000 & 1.000 \\
\bottomrule
\end{tabular}}
\end{table}

\subsection{What BrepEDIT retrieval measures}
With all same-part candidates retained, an instruction-independent scorer assigns one ranking per source. Since each edit is a target equally often, its aggregate score is at chance. Both the instruction-free head and the untrained source--candidate similarity score still attain $47.7$ MRR, under the reported scoring (which filters geometrically matching siblings for each query). Substituting an instruction from a different part also gives chance-level MRR. In the pooled gallery, the instruction-free head ranks the true outcome above all 100 edits of other parts in $98\%$ of queries. In \Cref{tab:s-brepedit-controls}, we list these protocols. Training and testing with a sibling's instruction gives $51.1$ MRR, $3.4$ points above chance.

A head that reads only the instruction and the candidate scores $64.6$ MRR on same-part retrieval, retaining $79\%$ of the full head's gain over chance. In the pooled gallery it falls to $27.0$ MRR, because the source is needed to identify the part. The task therefore depends heavily on whether the head can ground the instruction in the candidate's face geometry. The same head on raw face features scores $68.0$ MRR, compared with $69.1$ from \method and $67.8$ from same-corruption reconstruction. Against same-corruption reconstruction, the MRR gain persists across candidate counts ($0.7$--$2.0$ points) and is $1.0$ point when each query has exactly three candidates (\cref{tab:s-brepedit-difficulty}). MRR falls for every method as the number of candidates grows, from $79.8$ with three edits to $48.6$ with nine or more for \method, and \method has the highest score in every size group.

\begin{table}[htbp]
  \centering
  \caption{\textbf{Which inputs drive edit retrieval?} MRR and R@1 ($\times100$) for \method with same-part candidates or a pooled gallery that adds 100 edits of other parts. S, I, and C are source faces, instruction, and candidate faces. The upper block summarizes three encoder seeds. The next block uses the seed-42 encoder and three head seeds where standard deviations are shown. In the altered-instruction rows, $\mathrm{other}$ and $\mathrm{sib}$ indicate another part or another edit of the same part, respectively.}
  \label{tab:s-brepedit-controls}
  \footnotesize
  \setlength{\tabcolsep}{2.6pt}
  \begin{tabular}{@{}llcccc@{}}
    \toprule
    & & \multicolumn{2}{c}{Same part} & \multicolumn{2}{c}{Pooled} \\
    \cmidrule(lr){3-4}\cmidrule(l){5-6}
    Model & Reads & MRR & R@1 & MRR & R@1 \\
    \midrule
    Full head & S, I, C & $69.1{\pm}0.1$ & $50.4{\pm}0.2$ & $68.7{\pm}0.2$ & $49.9{\pm}0.2$ \\
    \quad another part's instruction at test & S, I$_{\mathrm{other}}$, C & $47.5{\pm}0.1$ & $22.3{\pm}0.1$ & -- & -- \\
    \midrule
    Full head, seed-42 encoder & S, I, C & $69.1{\pm}0.4$ & $50.4{\pm}0.5$ & $68.7{\pm}0.4$ & $50.0{\pm}0.5$ \\
    Trained without the instruction & S, C & $47.7{\pm}0.0$ & $22.4{\pm}0.0$ & $47.6{\pm}0.0$ & $22.4{\pm}0.0$ \\
    Trained without the source & I, C & $64.6{\pm}0.2$ & $44.4{\pm}0.3$ & $27.0{\pm}0.3$ & $12.9{\pm}0.2$ \\
    Trained and tested with a sibling's instruction & S, I$_{\mathrm{sib}}$, C & $51.1{\pm}0.2$ & $26.5{\pm}0.2$ & $50.4{\pm}0.0$ & $25.9{\pm}0.2$ \\
    Source--candidate face similarity (untrained) & S, C & $47.7$ & $22.5$ & $47.1$ & $22.4$ \\
    \midrule
    Chance &  & $47.7$ & $22.4$ & $5.0$ & $1.0$ \\
    \bottomrule
  \end{tabular}
\end{table}
\begin{table}[htbp]
  \centering
  \caption{\textbf{Edit retrieval at different candidate counts.} MRR and R@1 ($\times100$) on geometric instructions. The size columns group queries by the number of edits of the source part. For $n$ candidates after geometric-descriptor matching, normalized reciprocal rank is $(\mathrm{RR}-H_n/n)/(1-H_n/n)$, where $H_n$ is the $n$th harmonic number. The $k=3$ columns average over draws containing the true edit and two eligible siblings.}
  \label{tab:s-brepedit-difficulty}
  \footnotesize
  \setlength{\tabcolsep}{3pt}
  \begin{tabular}{@{}lcccccccccc@{}}
    \toprule
    & & \multicolumn{6}{c}{MRR by part size $n$} & & \multicolumn{2}{c}{$k{=}3$} \\
    \cmidrule(lr){3-8}\cmidrule(l){10-11}
    Method & MRR & $3$ & $4$ & $5$ & $6$ & $7$--$8$ & ${\geq}9$ & nMRR & MRR & R@1 \\
    \midrule
    Raw face features, no encoder & $68.0$ & $79.2$ & $71.2$ & $67.9$ & $65.2$ & $57.3$ & $46.9$ & $39.8$ & $78.4$ & $61.1$ \\
    \midrule
    Random initialization & $65.4{\pm}0.4$ & $77.3{\pm}0.2$ & $69.1{\pm}0.2$ & $64.3{\pm}0.4$ & $62.3{\pm}0.5$ & $54.2{\pm}0.9$ & $43.9{\pm}0.9$ & $34.9{\pm}0.7$ & $76.4{\pm}0.3$ & $57.7{\pm}0.6$ \\
    Same-corruption reconstruction & $67.8{\pm}0.0$ & $79.1{\pm}0.3$ & $71.2{\pm}0.1$ & $67.5{\pm}0.1$ & $65.4{\pm}0.4$ & $56.3{\pm}0.2$ & $46.7{\pm}0.1$ & $39.5{\pm}0.1$ & $78.3{\pm}0.1$ & $60.9{\pm}0.1$ \\
    Graph-JEPA objective & $63.9{\pm}0.7$ & $75.6{\pm}0.5$ & $67.5{\pm}0.6$ & $63.2{\pm}0.9$ & $61.7{\pm}0.4$ & $52.2{\pm}1.1$ & $42.5{\pm}0.6$ & $31.9{\pm}1.3$ & $75.2{\pm}0.5$ & $55.7{\pm}0.8$ \\
    \method & \hl $69.1{\pm}0.1$ & \hl $79.8{\pm}0.1$ & \hl $72.6{\pm}0.0$ & \hl $69.1{\pm}0.4$ & \hl $66.6{\pm}0.1$ & \hl $58.3{\pm}0.3$ & \hl $48.6{\pm}0.2$ & \hl $41.9{\pm}0.3$ & \hl $79.3{\pm}0.1$ & \hl $62.6{\pm}0.1$ \\
    Chance & $47.7$ & $61.2$ & $52.2$ & $45.7$ & $40.9$ & $35.9$ & $28.6$ & $0$ & $61.1$ & $33.3$ \\
    \midrule
    \method $-$ same-corr.\ recon. & $+1.3$ & $+0.7$ & $+1.4$ & $+1.6$ & $+1.2$ & $+2.0$ & $+1.9$ & $+2.4$ & $+1.0$ & $+1.7$ \\
    \method $-$ raw features & $+1.1$ & $+0.6$ & $+1.4$ & $+1.2$ & $+1.3$ & $+1.0$ & $+1.7$ & $+2.1$ & $+0.9$ & $+1.5$ \\
    \midrule
    Queries & $26{,}811$ & $7{,}029$ & $5{,}796$ & $4{,}380$ & $3{,}834$ & $3{,}396$ & $2{,}376$ & $26{,}811$ & $26{,}766$ & $26{,}766$ \\
    \bottomrule
  \end{tabular}
\end{table}

\subsection{Sensitivity to the coordinate frame}
We normalize the UV-grid positions in Fusion~360 and TMCAD data to a unit box. We use diagonal normalization for the NURBS control points. We repeat the probes with UV grids in the pretraining frame. On Fusion~360, the test score changes by less than one point for each of the three encoders tested at seed 42 (\method, the Graph-JEPA objective and random initialization). On TMCAD, \method changes from $80.08$ to $79.95$, while same-corruption reconstruction changes from $79.17$ to $80.01$, averaged over three seeds (\cref{tab:s-uvframe}). \method outperforms the Graph-JEPA objective by more than four points in both frames. Three TMCAD solids have degenerate bounding boxes and all-zero UV points in both settings.

\begin{table}[htbp]
  \centering
  \caption{\textbf{TMCAD probes in the deployed and pretraining coordinate frames.} Frozen-probe macro-F1 (\%) over three seeds. The deployed frame places UV points in a unit box. The pretraining frame uses bounding-box-diagonal normalization. Gaps are paired by seed with $95\%$ hierarchical intervals.}
  \label{tab:s-uvframe}
  \footnotesize
  \setlength{\tabcolsep}{5pt}
\begin{tabular}{lcccc}
\toprule
 & \multicolumn{2}{c}{Macro-F1 (\%)} & \multicolumn{2}{c}{\method $-$ method} \\
\cmidrule(lr){2-3}\cmidrule(l){4-5}
Method & Deployed & Pretraining & Deployed & Pretraining \\
\midrule
Random initialization (reference) & $77.33{\pm}0.95$ & $76.06{\pm}0.96$ & -- & -- \\
\midrule
Same-corruption reconstruction & $79.17{\pm}0.38$ & $80.01{\pm}0.23$ & $+0.91$\,{\scriptsize $[-0.72, +2.60]$} & $-0.06$\,{\scriptsize $[-1.63, +1.36]$} \\
Graph-JEPA objective & $75.00{\pm}1.71$ & $75.49{\pm}2.03$ & $+5.07$\,{\scriptsize $[+3.16, +7.27]$} & $+4.47$\,{\scriptsize $[+2.37, +6.78]$} \\
\midrule
\method & \hl $80.08{\pm}0.49$ & \hl $79.95{\pm}0.54$ & \hl -- & \hl -- \\
\bottomrule
\end{tabular}
\end{table}
\FloatBarrier

\section{Graph-JEPA optimization on the shared encoder}
\label{sec:s-graphjepa-schedules}

For the Graph-JEPA objective, we use Adam at $5{\cdot}10^{-4}$ with no weight decay, clipping, or warm-up. We use ReduceLROnPlateau to halve the learning rate with patience 20. The scheduler monitors the mean training loss every 79 updates. The learning rate threshold is $10^{-5}$, and the update cap of 30,000 is selected on validation at seed 42 from 2,400, 10,000, and 30,000 updates. The learning rate threshold stops the three runs at 27,571, 19,355, and 17,064 updates.

Additionally, we evaluate four alternative protocols: encoding target patches locally, retaining the official predictor and positional encoder configuration with a per-epoch EMA and 2,370 updates, using \method's optimizer and schedule with zero weight decay, and stepping the plateau scheduler once per ABC pass. The per-pass scheduler exhibits a sharp and persistent increase in training loss across all three seeds. We select the main-paper configuration from the first, fourth and fifth protocols using validation results at seed 42, before running the other two protocols. The shorter official configuration is more stable across seeds and has the highest three-seed validation mean among the Graph-JEPA adaptations. \method has a higher validation score on each benchmark than every adapted protocol at every tested seed (\cref{tab:s3-gj-adapt}).

\begin{table}[htbp]
  \centering
  \caption{\textbf{Graph-JEPA objective under five schedules on the shared encoder.} Frozen-probe macro-F1 (\%) over three pretraining seeds. The Mean column summarizes the per-seed validation means. The dagger marks the settings used in \mainref{Tables~1 and~2}.}
  \label{tab:s3-gj-adapt}
  \footnotesize
  \setlength{\tabcolsep}{5pt}
  \begin{tabular}{@{}lccccccc@{}}
  \toprule
  & \multicolumn{4}{c}{Validation} & \multicolumn{3}{c}{Test} \\
  \cmidrule(lr){2-5}\cmidrule(lr){6-8}
  Protocol & Fusion~360 & MFInstSeg & TMCAD & Mean & Fusion~360 & MFInstSeg & TMCAD \\
  \midrule
  original recipe, full-solid targets$^{\dagger}$ & 78.16 & 80.22 & 74.34 & $77.57{\pm}3.33$ & $78.68{\pm}3.15$ & $80.07{\pm}4.86$ & $75.00{\pm}1.71$ \\
  original recipe, patch-local targets & 75.07 & 74.44 & 72.24 & $73.91{\pm}9.65$ & $75.12{\pm}7.28$ & $74.16{\pm}15.42$ & $72.37{\pm}5.45$ \\
  official configuration (2{,}370 updates) & 78.55 & 80.59 & 76.14 & $78.43{\pm}0.13$ & $78.22{\pm}0.44$ & $80.45{\pm}0.57$ & $76.89{\pm}0.15$ \\
  \method's optimization (362.5k updates) & 65.19 & 67.42 & 73.05 & $68.55{\pm}1.66$ & $64.63{\pm}1.29$ & $67.20{\pm}3.18$ & $73.10{\pm}0.64$ \\
  plateau schedule per ABC pass & 66.73 & 66.26 & 67.57 & $66.85{\pm}1.59$ & $66.54{\pm}2.96$ & $66.11{\pm}3.86$ & $68.86{\pm}0.90$ \\
  \midrule
  \method & \hl \textbf{80.82} & \hl \textbf{89.60} & \hl \textbf{79.87} & \hl {\boldmath$83.43{\pm}0.16$} & \hl {\boldmath$81.29{\pm}0.39$} & \hl {\boldmath$89.61{\pm}0.32$} & \hl {\boldmath$80.08{\pm}0.49$} \\
  \bottomrule
  \end{tabular}
\end{table}

\end{document}